\pdfoutput=1
\documentclass[10pt,twocolumn,letterpaper]{article}
\usepackage[T1]{fontenc}
\usepackage[pagenumbers]{cvpr}

\usepackage{pifont}
\usepackage{colortbl}
\definecolor{oursrow}{RGB}{236,234,250}
\newcommand{\cmark}{{\color[RGB]{46,125,50}\ding{51}}}
\newcommand{\xmark}{{\color[RGB]{190,60,60}\ding{55}}}

\usepackage[expansion=false]{microtype}
\usepackage{placeins}
\usepackage{needspace}
\usepackage{amsmath,amssymb}
\definecolor{cvprblue}{rgb}{0.21,0.49,0.74}
\usepackage[pagebackref,breaklinks,colorlinks,allcolors=cvprblue]{hyperref}
\newcommand{\draftnote}[1]{}
\def\confName{CVPR}
\def\confYear{2027}
\title{InterMo: Streaming Interactive Co-Speech Motion Generation via Speech-Motion Interleaving}
\author{Jack Yu \quad Jiangtao Hu \quad Jamie \quad Leo Fung \quad Zhao Wang$^{\dagger *}$\\
LimX Dynamics\\
{\small $^{\dagger}$Project Lead \qquad $^{*}$Corresponding Author}\\
{\small Project page: \url{https://intermo-demo.pages.dev}}}
\begin{document}
\maketitle
\begin{abstract}
Interactive co-speech motion generation requires responding to both participants' speech and the preceding exchange while maintaining coherent motion as speech arrives. However, conditioning on separately encoded speech limits joint speech--motion context modeling, while training only on reference motion histories leaves generators unprepared for their own prediction errors. We present InterMo, a streaming framework that generates a target participant's upper-body motion from both participants' speech, supporting monologues and conversations. Speech--motion interleaving places both speech streams and the motion history in a time-aligned causal sequence, where shared self-attention jointly updates their representations at each streaming step and turn-taking cues preserve recent speaking history. Streaming-aligned training directly supervises decoded motion dynamics under mixed reference and predicted histories, with movement regularization against excessive motion and pose drift. On a 1,614\,h benchmark curated from five public datasets, InterMo achieves an overall score of 89.4 out of 100 across monologues and conversations, compared with 75.7 for the strongest of six retrained baselines, and most closely follows the ground truth in when to move and remain still during conversation. 
InterMo also remains stable over long sequences, generating each 80 ms step in 14.6 ms on a single GPU, and runs in real time on two humanoid robots without robot-specific training.
\end{abstract}

\begin{figure}[t]
\centering
\includegraphics[width=\linewidth]{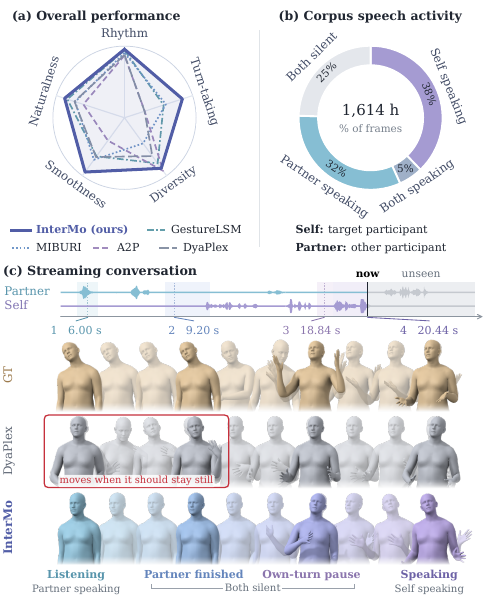}
\caption{\textbf{InterMo generates interactive motion as speech arrives.} (a) Over all evaluation clips, InterMo outperforms representative baselines, notably in Turn-taking. (b) Speech activity in our 1,614\,h corpus of monologue and conversation. (c) InterMo stays still until its own turn and gestures through its pause, like GT, whereas DyaPlex moves when it should stay still.}
\label{fig:teaser}
\end{figure}
\section{Introduction}
\label{sec:intro}

Co-speech motion reinforces speech rhythm and conveys meaning, making human communication more expressive~\cite{ginosar2019styles,kucherenko2020gesticulator,yoon2020trimodal}.
Recent advances in co-speech motion generation have enabled increasingly expressive and speech-aligned gestures in monologue settings~\cite{zhu2023diffgesture,chen2024diffsheg,liu2024emage,zhang2024semgesticulator,liu2025gesturelsm,zhang2025semtalk}.
Yet human communication is largely interactive, with speaking and listening interleaved through turn-taking~\cite{ekstedt2020turngpt,ekstedt2022vap}.
A participant's motion depends not only on their own speech but also on the partner's speech and the preceding exchange~\cite{ng2024audio2photoreal,nagano2026dyaplex}.
As this motion unfolds alongside the ongoing exchange, it must be generated continuously from incoming speech without access to future inputs, calling for streaming generation~\cite{mughal2026miburi,liu2026gesturefar,saleem2026livegesture}.
How to continuously update this interactive context while keeping the generated motion coherent remains an open problem.

Recent work has advanced both streaming motion generation and conversational motion modeling.
Streaming methods predict discrete motion codes~\cite{mughal2026miburi,saleem2026livegesture} or continuous latents~\cite{liu2026gesturefar} autoregressively from the observed speech and motion history.
MIBURI~\cite{mughal2026miburi} further couples such generation with LLM dialogue to drive interactive agents.
Conversational methods bring in the partner's speech through joint multimodal diffusion~\cite{mughal2024convofusion}, autoregressive guide-pose prediction followed by diffusion refinement~\cite{ng2024audio2photoreal}, or dyadic diffusion transformers~\cite{peng2026dyadit}, but generate offline from complete sequences.
DyaPlex~\cite{nagano2026dyaplex} goes further by generating causally from both participants.
However, these methods encode speech as a separate condition that the motion history cannot update as the conversation evolves. They are also trained only on reference motion histories with token- or latent-level objectives, which neither constrain the decoded motion dynamics nor expose the model to its own predictions during streaming~\cite{ross2011dagger,ranzato2016sequence}.

To address these limitations, we let the generated motion take part in forming the context for subsequent predictions. 
Incoming speech must be interpreted together with the preceding exchange and the motion already underway, which may need to continue, settle, or change as turns shift. We therefore allow both participants' speech and the target participant's motion history to refine one another's representations before predicting the next motion. Turn-taking cues complement this context by preserving recent speaking history, distinguishing conversational states that share the same current speech activity. The same formulation accommodates monologues with a silent partner. Since generated motion is fed back into this context, we train under histories that mix reference motion with the model's own predictions, and supervise and regularize the motion it decodes.

We present InterMo, a streaming framework that generates the target participant's upper-body motion from both participants' speech as the conversation unfolds.
To update this interactive context at each streaming step, we interleave self-speech, partner-speech, and motion-history tokens in a single sequence, allowing shared self-attention to jointly refine their representations.
Time-aligned rotary positions and a causal attention mask keep this sequence synchronized and free of future inputs.
A temporal--depth generator then predicts motion codes at each step, with turn-taking cues added to the motion tokens and modulating its temporal output.
Streaming-aligned training supervises the rotational increments of motion decoded through a frozen motion codec under mixed histories and penalizes excessive movement while constraining pose drift during reference holds.

Extensive evaluation is conducted on a 1,614\,h benchmark curated from five public datasets, covering both monologue and conversation settings, as summarized in Fig.~\ref{fig:teaser}(b). InterMo achieves the best overall score of 89.4 out of 100, compared with 75.7 for the strongest baseline, as shown in Fig.~\ref{fig:teaser}(a). It also most closely follows the ground truth in when to move and remain still during conversation, as illustrated in Fig.~\ref{fig:teaser}(c). Standard metrics on the official BEAT2 benchmark are reported for reference.
In continuous deployment, InterMo remains stable over long sequences while generating each 80 ms step in 14.6 ms on a single GPU. We also deploy the same generator on two humanoid robots via motion retargeting, enabling real-time execution without robot-specific training. Our main contributions are:
\begin{itemize}
    \item We present InterMo, a streaming interactive co-speech motion framework that generates a target participant's upper-body motion from both participants' speech, supporting monologues and conversations.
    \item We propose speech--motion interleaving, which jointly updates both speech streams and the motion history under time-aligned causal attention, complemented by turn-taking cues that preserve recent speaking history. 
    \item We develop streaming-aligned training, which mixes the model's own predictions into the motion history, supervises the dynamics of motion decoded under these histories, and regularizes excessive movement and pose drift.
    \item We achieve the best overall performance on held-out monologue and conversation clips from a corpus of 1,614\,h curated from five public datasets, and deploy InterMo in real time on two humanoid robots.
\end{itemize}

\begin{figure*}[t]
\centering
\includegraphics[width=\textwidth]{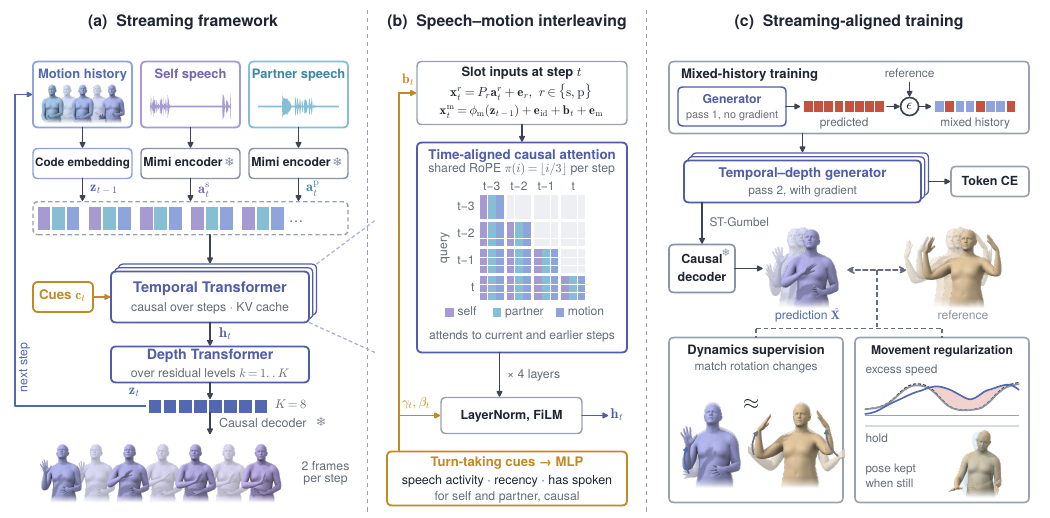}
\caption{\textbf{Overview of InterMo.} (a) The temporal--depth generator predicts motion codes from both speech streams and feeds them back as history. (b) Speech and motion slots update each other under time-aligned causal attention, with turn-taking cues conditioning the motion slot. (c) A gradient-free first pass builds mixed histories, under which the decoded motion is supervised and regularized.}
\label{fig:architecture}
\end{figure*}

\section{Related Work}
\label{sec:related}

\paragraph{Co-speech motion generation.}
Early co-speech motion methods learned speaker-specific gestures from video~\cite{ginosar2019styles,habibie2021video} and combined audio, text, and speaker identity~\cite{kucherenko2020gesticulator,yoon2020trimodal}. Subsequent work explored semantic and emotional cues~\cite{liu2022beat,zhang2024semgesticulator,chhatre2024amuse}, hierarchical audio--motion associations~\cite{liu2022ha2g}, diversity~\cite{li2021audio2gestures,qian2021templates,liu2024probtalk}, rhythm~\cite{ao2022rhythmic}, and style control~\cite{ghorbani2023zeroeggs,ao2023gesturediffuclip,yang2023diffusestylegesture}.
Diffusion methods advanced motion quality and temporal coherence~\cite{tevet2023mdm,zhu2023diffgesture,alexanderson2023listen,zhi2023livelyspeaker}. DiffSHEG jointly generates expressions and gestures~\cite{chen2024diffsheg}, GestureLSM models body-region dependencies with latent shortcut sampling~\cite{liu2025gesturelsm}, and SemTalk combines rhythmic motion with semantic emphasis~\cite{zhang2025semtalk}.
TalkSHOW, QPGesture, and EMAGE use discrete motion representations~\cite{yi2023talkshow,yang2023qpgesture,liu2024emage}, as do token-based text-to-motion models~\cite{zhang2023t2mgpt,guo2024momask,jiang2023motiongpt}.
These co-speech methods primarily target a single speaker and generate from the complete speech clip, whereas we condition motion causally on both participants' speech and accommodate monologues through a silent partner stream.

\paragraph{Conversational motion and dialogue modeling.}
Conversational motion generation conditions on partner speech through multimodal diffusion~\cite{mughal2024convofusion}, coarse-to-fine guide poses~\cite{ng2024audio2photoreal}, or dyadic diffusion transformers~\cite{peng2026dyadit}, generating from complete sequences. Other work addresses holistic speaker--listener motion~\cite{sun2024beyond}, text-guided two-person motion generation~\cite{liang2024intergen}, reactive two-person interaction~\cite{shi2024ittakestwo}, and facial listening and speaking--listening transitions~\cite{ng2022learning,zhou2022vico,tran2024dim,zhu2025infp,peng2025dualtalk}.
In spoken dialogue, full-duplex models place both speakers' streams in one sequence and learn turn-taking implicitly~\cite{nguyen2023dgslm,defossez2024moshi,veluri2024syncllm}, turn-taking models predict it explicitly from dialogue text~\cite{ekstedt2020turngpt} or from both participants' voice activity~\cite{ekstedt2022vap}, and SOLAMI places a character's speech and motion tokens in one sequence but alternates them turn by turn~\cite{jiang2025solami}.
DyaPlex conditions a causal motion tower on speech hidden states through cross-attention, using observed partner motion in its agent-only mode~\cite{nagano2026dyaplex}. We instead jointly update both speech streams and the target participant's motion history through interleaving, without requiring partner motion.

\paragraph{Streaming generation and exposure bias.}
Autoregressive motion models enable real-time character control~\cite{shi2024amdm,chen2024camdm,zhao2025dart,tevet2025closd} and streaming text-to-motion generation~\cite{xiao2025motionstreamer}.
For co-speech motion, DiffSHEG outpaints chunk by chunk~\cite{chen2024diffsheg}, RoboGesture generates one-second chunks with history-conditioned flow matching on a humanoid~\cite{wang2026robogesture}, GestureFAR predicts continuous latents token by token with a flow head~\cite{liu2026gesturefar}, and LiveGesture generates causal motion tokens per body region with zero look-ahead~\cite{saleem2026livegesture}.
MIBURI generates motion tokens causally alongside Moshi with a temporal--depth factorization but has no dedicated partner-speech input~\cite{mughal2026miburi,defossez2024moshi}. We adopt this factorization and focus on the temporal context and on training for rollout.
Models trained only on reference histories drift once their own predictions are fed back~\cite{ross2011dagger,ranzato2016sequence,li2017acrnn,lamb2016professor}.
Remedies expose training to imperfect histories, from history-token masking in LiveGesture~\cite{saleem2026livegesture} to noised or self-generated context in autoregressive video diffusion~\cite{chen2024diffusionforcing,huang2025selfforcing}. Unlike these remedies, which alter only the conditioning history, we also supervise the dynamics of motion decoded from discrete predictions under partially predicted histories, 
building on scheduled sampling~\cite{bengio2015scheduled}, straight-through Gumbel estimation~\cite{jang2016gumbel}, and rotational dynamics supervision~\cite{zhao2026kinest,wang2026timo}.

\section{InterMo}
\label{sec:method}
InterMo generates the target participant's upper-body motion from both participants' speech in a streaming manner, where its own generated motion takes part in updating the interactive context, as shown in Fig.~\ref{fig:architecture}.
Sec.~\ref{sec:preliminaries} introduces the streaming framework,
Sec.~\ref{sec:generator} builds this context with speech--motion interleaving and time-aligned causal attention, and Sec.~\ref{sec:kinematic} presents streaming-aligned training.

\subsection{Streaming Framework}
\label{sec:preliminaries}
\paragraph{Problem formulation.}
We refer to the target participant as self $s$ and the other participant as partner $p$; monologues use the same model with a null partner stream.
At each streaming step, the model conditions only on speech observed so far and previously generated motion.
The generated motion is represented by the local rotations of 13 SMPL-X~\cite{pavlakos2019smplx} upper-body joints, denoted as $\mathbf X=\{R_{n,j}\}$, where $R_{n,j}\in\mathrm{SO}(3)$ is the rotation of joint $j$ at motion frame $n$.
We focus on upper-body motion, as head, torso, and arm movements are more directly associated with speaking and listening than lower-body movements, and exclude fingers and face for consistency across datasets.

\paragraph{Speech and motion codecs.}
Speech and motion are represented as discrete tokens on a shared time grid at 12.5~Hz, so each streaming step corresponds to 80~ms.
Each speech stream is encoded by the pretrained Mimi codec~\cite{defossez2024moshi}, and the token embeddings across its $Q$ codebooks are summed to obtain audio features $\mathbf a_t^{\mathrm{s}}$ and $\mathbf a_t^{\mathrm{p}}$ at step $t$.
Motion is tokenized by a causal residual vector-quantized VAE (RVQ-VAE)~\cite{oord2017vqvae,lee2022rq}, trained on our motion corpus and frozen during generator training.
Its encoder maps 6D joint rotations~\cite{zhou2019continuity} into $K$ residual indices $\mathbf z_t=(z_{t,1},\ldots,z_{t,K})$ at each step, covering two motion frames at 25fps.
The causal decoder $\mathcal D$ reconstructs motion from the corresponding summed codebook embeddings.

\paragraph{Framework.}
Following temporal--depth modeling~\cite{defossez2024moshi,mughal2026miburi}, a temporal Transformer captures dependencies across steps and yields the context feature $\mathbf h_t$, which predicts the first motion code $z_{t,1}$ and conditions a depth Transformer that autoregressively predicts the remaining codes $z_{t,2:K}$ from the preceding codes within the same step and the current audio features.
Motion tokens are thus factorized as
\begin{equation}
 p_\theta(\mathbf z_{1:T}\mid\mathcal C)
 =\prod_{t=1}^{T}\prod_{k=1}^{K}
 p_\theta\!\left(z_{t,k}\mid
 \mathbf z_{<t},\mathbf z_{t,<k},\mathcal C_{\leq t}\right),
\label{eq:factorization}
\end{equation}
where $\mathcal C_{\leq t}$ comprises both speech streams and the turn-taking cues of Sec.~\ref{sec:generator}, all derived from observed speech up to step $t$, together with a motion-style embedding indexed by data source.
Handling residual levels within each step keeps the temporal sequence length independent of $K$.
During streaming, the temporal Transformer processes each new step incrementally with its key--value cache, which stays constant in size under the bounded attention windows of Sec.~\ref{sec:generator}. The predicted codes are decoded causally into motion frames and fed back as motion history for the next streaming step. Speech and turn-taking conditions are randomly dropped during training, and at inference classifier-free guidance~\cite{ho2022classifier} contrasts the full condition with null speech tokens while keeping the cues.

\subsection{Speech--Motion Interleaving}
\label{sec:generator}
We build the interactive context by interleaving both speech streams with motion history, allowing incoming speech to be represented in the context of the preceding exchange and ongoing motion. Time-aligned causal attention synchronizes the three streams without accessing future inputs, while turn-taking cues preserve recent speaking history.

\paragraph{Interleaved streams.}
At each step $t$, the self-speech, partner-speech, and motion streams each add one slot:
\begin{equation}
\begin{aligned}
 \mathbf x_t^{r} &= P_{r}\mathbf a_t^{r}+\mathbf e_{r},\qquad r\in\{\mathrm{s},\mathrm{p}\},\\
 \mathbf x_t^{\mathrm{m}} &= \phi_{\mathrm{m}}(\mathbf z_{t-1})+\mathbf e_{\mathrm{id}}+\mathbf b_t+\mathbf e_{\mathrm{m}},
\end{aligned}
\label{eq:interleaved_inputs}
\end{equation}
where $P_{r}$ are projections, $\mathbf e_{\mathrm{s}},\mathbf e_{\mathrm{p}},\mathbf e_{\mathrm{m}}$ are learned role embeddings, $\phi_{\mathrm{m}}$ sums the embeddings of the previous step's $K$ codes (a start token at $t=1$), $\mathbf e_{\mathrm{id}}$ is a motion-style embedding indexed by data source, and $\mathbf b_t$ is the turn-taking embedding defined below.
The temporal Transformer processes the interleaved sequence $[\mathbf x_1^{\mathrm{s}},\mathbf x_1^{\mathrm{p}},\mathbf x_1^{\mathrm{m}},\mathbf x_2^{\mathrm{s}},\ldots]$.
At each layer, the current speech slots attend to motion history and to one another, while the motion slot integrates both speech streams. This lets motion history influence the speech representations that contribute to the next prediction, a path absent from one-way cross-attention in which motion queries read separately encoded speech features.

\paragraph{Time-aligned causal attention.}
Since the three slots of a step share the same time, we assign them a shared rotary position~\cite{su2021roformer}, $\pi(i)=\lfloor i/3\rfloor$ for zero-based sequence index $i$, and leave the role embeddings to identify the streams.
Attention follows a stream-specific causal mask in Fig.~\ref{fig:architecture}(b),
\begin{equation}
 M_{ij}=\begin{cases}
 0,&0\leq\pi(i)-\pi(j)<L_{\sigma(j)},\\
 -\infty,&\text{otherwise},
 \end{cases}
\label{eq:stream_mask}
\end{equation}
where $\sigma(j)$ is the stream of key $j$ and $L_{\sigma(j)}$ its visible window in streaming steps.
Attention is thus causal across steps and bidirectional within a step, without future leakage since the motion slot holds only the previous step's codes.

\paragraph{Turn-taking cues.}
A short pause within one's own turn and a long silence after the partner has finished share the same current speech activity, and the speech that separates them can fall outside the attention window.
To make recent speaking history explicit, we compute for each participant $r\in\{\mathrm{s},\mathrm{p}\}$, causally from voice activity, the activity bit $v_t^r$, the elapsed time $d_t^r$ since $r$ last spoke, and a flag $\kappa_t^r$ indicating whether $r$ has spoken yet.
The elapsed time is capped at $d_{\max}$ and converted to a normalized recency $u_t^r=\log(1+\min(d_t^r,d_{\max}))/\log(1+d_{\max})$, which is 0 during speech and set to 1 before $r$ has spoken.
We define the cues as $\mathbf c_t=[\eta_t,v_t^{\mathrm{s}},v_t^{\mathrm{p}},u_t^{\mathrm{s}},u_t^{\mathrm{p}},\kappa_t^{\mathrm{s}},\kappa_t^{\mathrm{p}}]$, where $\eta_t$ is 0 when the cues are dropped during training. Voice activity comes from reference labels during training and evaluation and from a causal energy-based detector at deployment.
An MLP maps $\mathbf c_t$ to the turn-taking embedding $\mathbf b_t$ in Eq.~\eqref{eq:interleaved_inputs} and to a feature-wise scale and shift~\cite{perez2018film} on the normalized temporal motion output, which gives the context feature $\mathbf h_t$.

\begin{figure}[t]
\centering
\includegraphics[width=\linewidth]{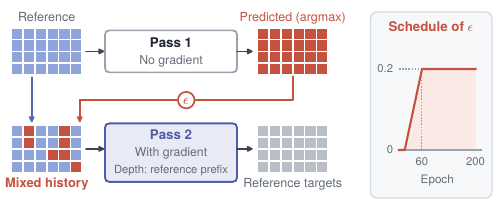}
\caption{\textbf{Mixed-history training.} Pass~1 predicts codes without gradient; each history code is replaced by its prediction with probability $\epsilon$, and pass~2 trains on this mixed history.}
\label{fig:mixed_history}
\end{figure}

\subsection{Streaming-Aligned Training}
\label{sec:kinematic}
Since each motion prediction becomes part of the context for subsequent steps, we supervise decoded motion under partially predicted histories. Mixed-history training introduces the model's predictions into the reference motion history; dynamics supervision matches the rotational increments of motion decoded under these histories; and movement regularization penalizes excessive movement and anchors poses during holds. The overall objective is $\mathcal L=\mathcal L_{\mathrm{CE}}+\lambda_{\mathrm{dyn}}\mathcal L_{\mathrm{dyn}}+\lambda_{\mathrm{reg}}\mathcal L_{\mathrm{reg}}$, with token cross-entropy $\mathcal L_{\mathrm{CE}}$.

\paragraph{Mixed-history training.}
We construct partially predicted motion histories using a two-pass procedure, as shown in Fig.~\ref{fig:mixed_history}. Scheduled sampling~\cite{bengio2015scheduled} and its use in autoregressive motion models~\cite{li2017acrnn} expose a model to its own predictions; we additionally supervise decoded motion dynamics, with both token and motion losses conditioned on the same mixed histories.
A first pass under teacher forcing without gradients yields argmax codes $\hat z_{t,k}$, and each code of the motion history is replaced by the prediction with probability $\epsilon$, i.e., $\tilde z_{t,k}=\hat z_{t,k}$ if $m_{t,k}=1$ and $z_{t,k}$ otherwise, with $m_{t,k}\sim\mathrm{Bernoulli}(\epsilon)$, where $\epsilon$ ramps linearly to $\epsilon_{\max}$ after an initial teacher-forcing stage.
A second pass with gradients places $\tilde{\mathbf z}_{t-1}$ in the motion slot of step $t$ and keeps reference targets, and the token cross-entropy $\mathcal L_{\mathrm{CE}}$ and the motion losses below are all computed from this pass.
Within each step, the depth Transformer remains conditioned on the reference codes of the preceding levels, as finer residual codes are unreliable once a coarser code is wrong.
To decode motion from this pass, straight-through Gumbel sampling~\cite{jang2016gumbel} selects hard RVQ codes in the forward pass and backpropagates through a soft relaxation, and the frozen decoder $\mathcal D$ maps the summed codebook embeddings to the predicted motion $\hat{\mathbf X}$, so motion losses reach the generator without updating the codec.

\paragraph{Dynamics supervision.}
Since many poses are plausible for the same speech, our primary motion supervision targets dynamics rather than absolute poses.
We therefore use the rotational increments $\boldsymbol\omega_{n,j}=\operatorname{Log}(R_{n,j}^{\top}R_{n+1,j})$ and their temporal differences $\boldsymbol\alpha_{n,j}=\boldsymbol\omega_{n+1,j}-\boldsymbol\omega_{n,j}$, as in~\cite{zhao2026kinest,wang2026timo},
\begin{equation}
 \mathcal L_{\mathrm{dyn}}=\sum_{f\in\mathcal F}\lambda_f\big(
 \langle\rho_\beta(\hat{\boldsymbol\omega}-\boldsymbol\omega)\rangle_f
 +\langle\rho_\beta(\hat{\boldsymbol\alpha}-\boldsymbol\alpha)\rangle_f\big),
\label{eq:dynamics_losses}
\end{equation}
where $\rho_\beta$ is Smooth-L1, $\langle\cdot\rangle_{\mathcal A}$ denotes the mean over set $\mathcal A$, and $\mathcal F$ comprises all frames, low-motion frames of the reference, and frames where neither participant speaks, weighted by $\lambda_f$.
Each term is averaged within its own frame set, so rare restrained motion is not diluted by active intervals. As $\boldsymbol\omega$ and $\boldsymbol\alpha$ are per-joint angular velocity and acceleration, this loss constrains how motion evolves rather than which pose is reached, so constant pose offsets are not penalized.

\paragraph{Movement regularization.}
We complement frame-level dynamics matching with a window-level penalty on excessive movement and an absolute-pose constraint during reference holds.
Let $\nu_{w,j}$ denote the per-frame angular displacement $\|\boldsymbol\omega_{n,j}\|$ of arm joint $j$ averaged over frames $n$ in window $w$, with $\hat\nu_{w,j}$ its counterpart in the predicted motion.
$\mathcal L_{\mathrm{reg}}$ combines a one-sided speed term that only penalizes excess movement and a hold term that anchors all generated joints to the reference pose during holds to limit drift,
\begin{equation}
\begin{aligned}
 \mathcal L_{\mathrm{reg}}={}&
 \lambda_{\mathrm{spd}}\big\langle[\hat\nu-\gamma\nu]_+\big\rangle_{\mathcal W}\\
 &+\lambda_{\mathrm{hold}}\big\langle\rho_\beta\big(\operatorname{Log}(\hat R^{\top}R)\big)\big\rangle_{\mathcal S},
\end{aligned}
\label{eq:reg_losses}
\end{equation}
where $[\cdot]_+=\max(\cdot,0)$, $\gamma$ is a multiplicative margin, $\lambda_{\mathrm{spd}}$ and $\lambda_{\mathrm{hold}}$ weight the two terms, $\mathcal W$ comprises all windows and arm joints, and $\mathcal S$ comprises frames where the reference arms are nearly still, averaged over all joints.
Window length, thresholds, and weights are given in the supplement.

\section{Experiments}
\label{sec:experiments}

\begin{table*}[t]
\centering
\caption{\textbf{Comparison with state-of-the-art methods.} Scores range from 0 to 100 and measure agreement with GT reference statistics; higher is better. Avg. denotes the mean across dimensions; Turn-taking is evaluated only on conversation clips. Partner indicates access to partner speech; Online indicates incremental generation as speech arrives. Bold: best; underlined: second best.}
\label{tab:main}
\small
\setlength{\tabcolsep}{4.6pt}
\begin{tabular}{lccccccccccccc c}
\toprule
& & & \multicolumn{5}{c}{Monologue} & \multicolumn{6}{c}{Conversation} & All\\
\cmidrule(lr){4-8}\cmidrule(lr){9-14}\cmidrule(lr){15-15}
Method & Partner & Online & Rhy. & Nat. & Smo. & Div. & Avg. & Rhy. & Nat. & Smo. & Div. & Turn. & Avg. & Avg.\\
\midrule
SemTalk~\cite{zhang2025semtalk} & \xmark & \xmark & 48.6 & 26.8 & 7.4 & 66.9 & 37.4 & 52.5 & 37.2 & 7.9 & 74.7 & 35.0 & 41.5 & 41.0\\
GestureLSM~\cite{liu2025gesturelsm} & \xmark & \xmark & \underline{98.0} & 78.1 & 66.6 & 85.4 & \underline{82.0} & \underline{89.8} & \underline{82.0} & 66.0 & 75.2 & 57.1 & \underline{74.0} & \underline{75.7}\\
DiffSHEG~\cite{chen2024diffsheg} & \xmark & \cmark & 69.2 & \underline{79.8} & 24.0 & 74.6 & 61.9 & 82.7 & 78.8 & 24.6 & 69.5 & \underline{77.1} & 66.5 & 66.1\\
MIBURI~\cite{mughal2026miburi} & \xmark & \cmark & 94.4 & 75.0 & \underline{80.7} & 54.9 & 76.2 & 86.4 & 65.4 & \underline{68.0} & 39.8 & 60.3 & 64.0 & 67.7\\
Audio2Photoreal~\cite{ng2024audio2photoreal} & \cmark & \xmark & 83.2 & 57.8 & 39.4 & \underline{87.8} & 67.0 & \underline{89.8} & 57.4 & 39.1 & \textbf{89.7} & 29.5 & 61.1 & 61.3\\
DyaPlex~\cite{nagano2026dyaplex} & \cmark & \cmark & 91.9 & 63.1 & 71.2 & 68.6 & 73.7 & 83.3 & 71.2 & 65.8 & 64.5 & 29.3 & 62.8 & 64.6\\
\midrule
\rowcolor{oursrow}InterMo (ours) & \cmark & \cmark & \textbf{98.5} & \textbf{84.2} & \textbf{87.8} & \textbf{89.1} & \textbf{89.9} & \textbf{90.1} & \textbf{84.8} & \textbf{93.7} & \underline{82.2} & \textbf{84.5} & \textbf{87.0} & \textbf{89.4}\\
\bottomrule
\end{tabular}
\end{table*}

We evaluate overall motion in Sec.~\ref{sec:quality}, interactive motion in Sec.~\ref{sec:turns}, each component's contribution in Sec.~\ref{sec:ablation}, and continuous deployment in Sec.~\ref{sec:applications}.

\subsection{Setup}\label{sec:setup}
\paragraph{Dataset.}
We train and evaluate on a benchmark that we build from five public datasets.
It contains three conversational sources, namely Seamless Interaction~\cite{agrawal2025seamless}, Embody 3D~\cite{mclean2025embody}, and Audio2Photoreal~\cite{ng2024audio2photoreal}, and two monologue sources, BEAT2~\cite{liu2024emage} and Trinity~\cite{ferstl2018trinity}.
We convert all motion to 13 upper-body joints at 25\,fps and filter out segments with reconstruction artifacts, locomotion, or large motion during mutual silence, yielding a 1,614\,h corpus.
For evaluation, we sample 200 held-out clips of up to 90\,s and group them by whether the partner speaks into 64 monologue and 136 conversation clips.
A paired bootstrap over evaluation clips supports InterMo's average-score advantage over each baseline on the monologue subset, the conversation subset, and the full evaluation set.
Data sources, processing, splits, and the bootstrap analysis are detailed in the supplement.

\paragraph{Metrics.}
We evaluate five dimensions using the ground-truth (GT) motion of the same clips as reference. Since reading jerk as lower-is-better or variation as higher-is-better would favor overly still or exaggerated motion, each metric is scored by its agreement with GT on a common 0--100 scale. \textit{Rhythm} measures the agreement of speech--motion beat consistency (BC) with GT on self-speech frames. \textit{Naturalness} uses Fr\'echet gesture distance (FGD) in a learned motion latent space~\cite{peng2026dyadit}. \textit{Smoothness} measures agreement with GT jerk statistics~\cite{wang2026timo}, accounting for both overly damped and jittery motion. \textit{Diversity} measures agreement with GT within-clip motion variation (L1div)~\cite{mughal2026miburi}. \textit{Turn-taking}, evaluated on conversation clips, combines listening motion activity and whether motion beats after partner pauses follow GT timing beyond chance. The average score is the arithmetic mean of the dimension scores: four for monologues and five for conversations and the overall evaluation. Scoring formulas and raw values are in the supplement; Tab.~\ref{tab:beat2} uses the official BEAT2 split and metrics.

\paragraph{Baselines.}
We compare with the single-speaker methods DiffSHEG~\cite{chen2024diffsheg}, SemTalk~\cite{zhang2025semtalk}, GestureLSM~\cite{liu2025gesturelsm}, and MIBURI~\cite{mughal2026miburi}, and the conversational methods Audio2Photoreal~\cite{ng2024audio2photoreal} and DyaPlex~\cite{nagano2026dyaplex}. All methods are retrained on our curated corpus with the common 13-joint upper-body representation. We use released implementations where available and reimplement DyaPlex from its published description. Single-speaker methods receive self-speech as in their original design, so all methods share the same input on monologue clips. Audio2Photoreal, DyaPlex, and InterMo also receive partner speech, and DyaPlex additionally receives GT partner motion on conversation clips, which InterMo does not use. Details are in the supplement.

\begin{figure*}[t]
\centering
\includegraphics[width=\textwidth]{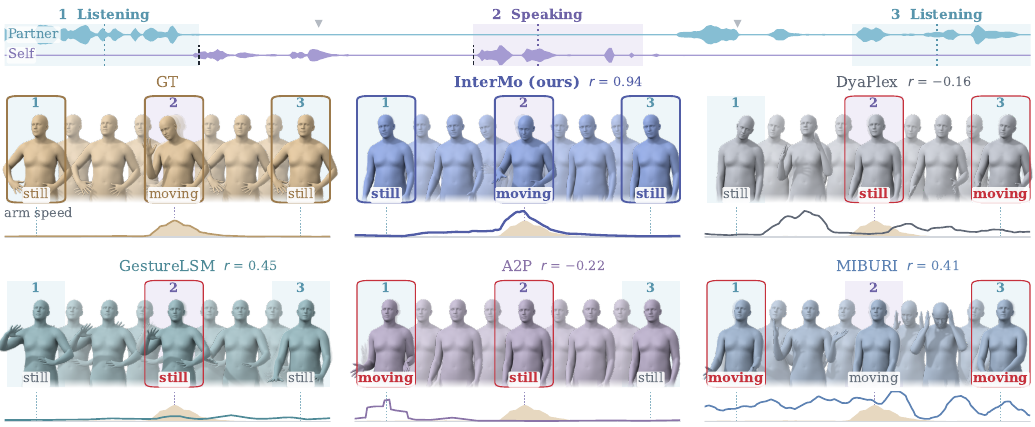}
\caption{\textbf{Qualitative comparison with state-of-the-art methods.} InterMo starts gesturing when the target participant takes the turn and settles when the partner speaks again. Its movement over time correlates best with GT at $r{=}0.94$. Red boxes mark mismatches with GT.}
\label{fig:qualitative}
\end{figure*}

\subsection{Quantitative Results}
\label{sec:quality}
We compare InterMo with retrained baselines on our held-out evaluation clips. Fig.~\ref{fig:teaser}(a) summarizes performance on the full evaluation set, where InterMo scores 89.4 averaged across five dimensions, compared with 75.7 for the strongest baseline. Tab.~\ref{tab:main} further reports results separately for monologue and conversation subsets, with InterMo leading in both. We also train our model on the official BEAT2 training split and report test performance in Tab.~\ref{tab:beat2}.

\begin{table}[!t]
\centering
\caption{\textbf{Comparison of upper-body motion on BEAT2.} $\downarrow$: lower is better;  $\rightarrow$: closer to GT is better; jerk in $10^{3}$. Bold: best; underlined: second best.}
\label{tab:beat2}
\small
\setlength{\tabcolsep}{4.2pt}
\begin{tabular}{lccccc}
\toprule
Method & Online & FGD$\downarrow$ & BC$\rightarrow$ & L1div$\rightarrow$ & Jerk$\rightarrow$\\
\midrule
{\color{gray}GT} & {\color{gray}--} & {\color{gray}--} & {\color{gray}0.554} & {\color{gray}0.636} & {\color{gray}0.37}\\
\midrule
SemTalk~\cite{zhang2025semtalk} & \xmark & \underline{0.090} & \underline{0.582} & \underline{0.660} & 0.14\\
GestureLSM~\cite{liu2025gesturelsm} & \xmark & 0.116 & \textbf{0.577} & 0.670 & \underline{0.27}\\
DiffSHEG~\cite{chen2024diffsheg} & \cmark & 0.128 & 0.460 & 0.694 & 0.17\\
MIBURI~\cite{mughal2026miburi} & \cmark & 0.318 & 0.719 & 1.277 & 1.38\\
\midrule
\rowcolor{oursrow}InterMo (ours) & \cmark & \textbf{0.065} & 0.612 & \textbf{0.655} & \textbf{0.32}\\
\bottomrule
\end{tabular}
\end{table}

\paragraph{Conversation.}
InterMo obtains an average score of 87.0, compared with 74.0 for the strongest baseline. It leads in Rhythm and Naturalness, with the largest advantage in Smoothness, scoring 93.7 while no baseline exceeds 68.0. Audio2Photoreal achieves the highest Diversity of 89.7, but its lower Naturalness and Smoothness scores of 57.4 and 39.1 indicate that matching within-clip variation alone does not ensure comparable motion quality across dimensions. The raw metrics in the supplement show that Audio2Photoreal and DiffSHEG produce overly damped motion, whereas DyaPlex and MIBURI exhibit excessive jerk. Fig.~\ref{fig:qualitative} further illustrates InterMo's closer agreement with GT in temporal movement patterns. 

\paragraph{Monologue.}
InterMo maintains its lead with an average score of 89.9, ahead of GestureLSM at 82.0, and is closest to GT on all four dimensions. This advantage holds even though InterMo generates strictly causally and some baselines have access to the complete utterance. Together with the conversation results, these findings demonstrate that the same framework supports both settings by treating a monologue as a conversation with a silent partner.

\paragraph{BEAT2 benchmark.}
To evaluate performance beyond our curated benchmark and scoring protocol, we train InterMo and the four single-speaker baselines on the official BEAT2~\cite{liu2024emage} English training split and evaluate on its official test split. All methods use an audio-only setting, generate without GT seed frames, and are evaluated on the same 13 upper-body joints. As shown in Tab.~\ref{tab:beat2}, InterMo attains the lowest FGD of 0.065 and is closest to GT in jerk and L1div. These results provide additional evidence of InterMo's motion quality under standard metrics.

\begin{figure}[!t]
\centering
\includegraphics[width=\linewidth]{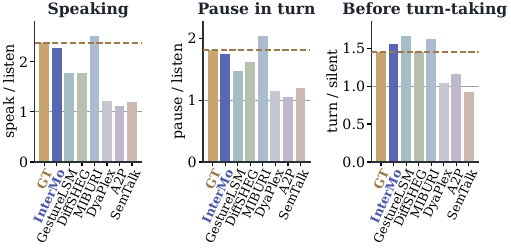}
\caption{\textbf{Motion across conversational states.} Relative arm speed on the conversation clips. Left, middle: speaking and own-turn pauses relative to listening. Right: after the partner stops, arm speed before taking the turn within 2 s relative to staying silent.}
\label{fig:states}
\end{figure}

\subsection{Interactive Motion Quality}
\label{sec:turns}

\paragraph{Listening and response timing.}
InterMo ranks first in Turn-taking with a score of 84.5 in Tab.~\ref{tab:main}.
DiffSHEG ranks second at 77.1 using only self-speech; its damped motion remains limited during self-silence and follows the resumption of self-speech.
Audio2Photoreal and DyaPlex hear the partner yet score only 29.5 and 29.3; their listening movements are $1.8\times$ and $3.5\times$ the GT level, respectively, and their motion-beat timing after partner pauses is at chance level.
These results show that partner speech alone does not ensure appropriate listening activity and motion timing.

\paragraph{Conversational states.}
We label conversation frames as speaking, listening, own-turn pauses, or silence right after the partner stops, as defined in the supplement. The left and middle panels of Fig.~\ref{fig:states} compare arm speed during speaking and own-turn pauses relative to listening. GT arm speed is $2.4\times$ the listening level during speech and $1.8\times$ during own-turn pauses, closely matched by InterMo at $2.3\times$ and $1.8\times$, respectively. GestureLSM and DiffSHEG reach $1.8\times$ the listening level during speech, and Audio2Photoreal, SemTalk, and DyaPlex at most $1.2\times$. MIBURI overshoots at $2.5\times$ during speech and $2.0\times$ during own-turn pauses. InterMo thus keeps gesturing through own-turn pauses and settles while listening, as Fig.~\ref{fig:qualitative} also illustrates.

\paragraph{Before taking the turn.}
Right after the partner stops, GT arm speed is $1.5\times$ as high when the target participant takes the turn within the next 2\,s as when they stay silent, as shown in the right panel of Fig.~\ref{fig:states}.
InterMo reproduces this ratio at $1.6\times$, whereas Audio2Photoreal and DyaPlex, which also hear the partner, change little.
MIBURI also speeds up, but it overshoots every contrast in Fig.~\ref{fig:states} and moves excessively throughout the exchange in Fig.~\ref{fig:qualitative}.
These results suggest that InterMo captures differences in motion activity that are associated with subsequent turn-taking.

\begin{table}[!t]
\centering
\caption{\textbf{Ablation study.} Higher scores indicate closer agreement with GT. Avg. is the mean across the five dimensions.}
\label{tab:ablation}
\small
\setlength{\tabcolsep}{3.6pt}
\begin{tabular}{lcccccc}
\toprule
Configuration & Rhy. & Nat. & Smo. & Div. & Turn. & Avg.\\
\midrule
\rowcolor{oursrow}InterMo (full) & \textbf{95.1} & \textbf{87.4} & \textbf{93.2} & \textbf{86.8} & \textbf{84.5} & \textbf{89.4}\\
\midrule
\multicolumn{7}{l}{\textit{Interactive context}}\\
\quad w/o partner speech & 93.8 & 84.5 & 90.2 & 71.2 & 77.6 & 83.5\\
\quad w/o interleaving & 91.2 & 86.8 & 84.6 & 76.5 & 70.0 & 81.8\\
\quad w/o turn-taking cues & 92.9 & 83.7 & 86.1 & 71.9 & 71.2 & 81.2\\
\midrule
\multicolumn{7}{l}{\textit{Streaming-aligned training}}\\
\quad w/o dynamics & 94.0 & 86.3 & 91.9 & 70.6 & 81.5 & 84.9\\
\quad w/o regularization & 93.1 & 86.8 & 84.7 & 61.1 & 73.7 & 79.9\\
\quad w/o mixed history & 90.0 & 82.8 & 75.0 & 61.1 & 41.3 & 70.0\\
\bottomrule
\end{tabular}
\end{table}

\subsection{Ablation Study}
\label{sec:ablation}
\paragraph{Interactive context.}
As shown in Tab.~\ref{tab:ablation}, removing partner speech, interleaving, or turn-taking cues lowers the overall score from 89.4 to 83.5, 81.8, and 81.2, respectively. 
Without interleaving, motion slots read both speech streams through separate causal cross-attention with identical inputs and windows, but speech features are no longer updated by motion history or by each other. 
All three variants lose most in Turn-taking and Diversity, with excessive motion variation; removing interleaving or turn-taking cues also increases jitter. These results support the contribution of the interactive context to motion quality and to the listening activity and timing analyzed in Sec.~\ref{sec:turns}.

\paragraph{Streaming-aligned training.}
Without dynamics supervision, Diversity decreases while Smoothness remains largely unchanged, suggesting that rotational-increment supervision also helps control motion amplitude. Omitting movement regularization leads to excessive and jittery motion. Training only on reference histories degrades every dimension, with Turn-taking falling from 84.5 to 41.3, as the model never sees its own predictions during training. These ablations indicate that the interactive context improves prediction from both speech streams and motion history, while streaming-aligned training keeps the motion fed back into that context close to GT, limiting error accumulation.

\begin{figure}[!t]
\centering
\includegraphics[width=\linewidth]{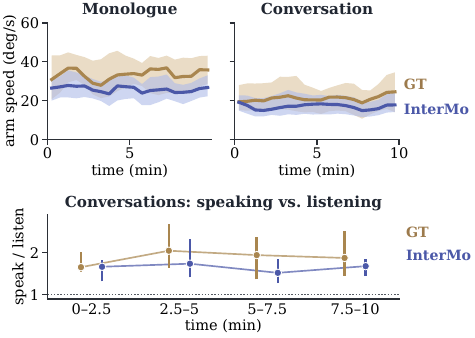}
\caption{\textbf{Long-sequence generation.} Top: median arm speed of GT and InterMo. Bottom: speaking-to-listening speed ratio per 2.5\,min of conversation. Shading and bars: inter-quartile range.}
\label{fig:longseq}
\end{figure}

\begin{figure}[t]
\centering
\includegraphics[width=\linewidth]{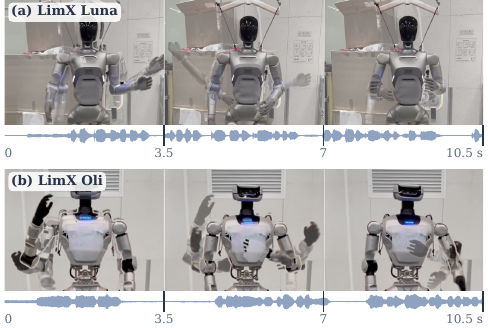}
\caption{\textbf{Real-time humanoid execution.} (a) and (b) show InterMo driving LimX Luna and LimX Oli, respectively. Each image overlays poses within a 3.5\,s interval, with lighter poses indicating earlier times. Waveforms show input speech.}
\label{fig:robot_execution}
\end{figure}

\subsection{Continuous Real-Time Deployment}
\label{sec:applications}
\paragraph{Real-time streaming.}
InterMo takes an average of 14.6\,ms to generate each 80\,ms step on a single GPU, enabling real-time streaming. Computing the turn-taking cues with the causal detector instead of reference labels yields an overall score of 89.0, compared with 89.4, so InterMo does not depend on reference labels.
To assess continuous operation, we stream the first 10 minutes of 33 held-out recordings through it.
As Fig.~\ref{fig:longseq} shows, arm speed follows the temporal trend of GT, while remaining higher during speaking than listening in each 2.5-minute interval of the conversations.
In a live conversational demo, InterMo continuously drives an avatar from the speech of the user and a dialogue agent, computing the turn-taking cues with the same causal detector; the setup and examples are in the supplement.

\paragraph{Humanoid execution.}
We deploy InterMo on LimX Luna~\cite{limx_luna} and LimX Oli~\cite{limx_oli}, retargeting its real-time output to each robot and tracking it with a whole-body controller~\cite{fu2026cosa05}.
In Fig.~\ref{fig:robot_execution}, both robots gesture with speech using the same generator without robot-specific training. 

\FloatBarrier

\section{Conclusion}
\label{sec:conclusion}
We presented InterMo, a streaming framework for interactive co-speech motion generation in monologues and conversations. Speech--motion interleaving jointly updates speech and motion representations, while turn-taking cues preserve recent speaking history. Since generated motion feeds back into this context, streaming-aligned training supervises decoded motion dynamics under mixed reference and predicted histories and regularizes excessive movement and pose drift. 
Evaluation on our curated 1,614\,h benchmark shows that InterMo outperforms six retrained baselines overall and most closely follows GT in when to move and remain still during conversation. 
Deployment experiments further demonstrate continuous real-time generation over 10-minute streams and execution on two humanoid robots without robot-specific training. The current framework focuses on upper-body motion, does not incorporate the partner's visual cues, and relies on estimated speech activity for turn-taking cues. Future work will explore richer multimodal context and generate hand and facial motion.

\section*{LLM Usage Statement}
ChatGPT and Claude were used for language polishing and grammar correction in parts of the manuscript.
All scientific content, methodology, experiments, and conclusions were developed, checked, and verified by the authors.

\FloatBarrier
{\small
\bibliographystyle{ieeenat_fullname}
\bibliography{main}

\begin{thebibliography}{78}
\providecommand{\natexlab}[1]{#1}
\providecommand{\url}[1]{\texttt{#1}}
\expandafter\ifx\csname urlstyle\endcsname\relax
  \providecommand{\doi}[1]{doi: #1}\else
  \providecommand{\doi}{doi: \begingroup \urlstyle{rm}\Url}\fi

\bibitem[Agrawal et~al.(2025)Agrawal, Akinyemi, Alvero, Behrooz, Buffalini,
  Carlucci, Chen, Chen, Chen, Cheng, et~al.]{agrawal2025seamless}
Vasu Agrawal, Akinniyi Akinyemi, Kathryn Alvero, Morteza Behrooz, Julia
  Buffalini, Fabio~Maria Carlucci, Joy Chen, Junming Chen, Zhang Chen, Shiyang
  Cheng, et~al.
\newblock Seamless interaction: Dyadic audiovisual motion modeling and
  large-scale dataset.
\newblock \emph{arXiv preprint arXiv:2506.22554}, 2025.

\bibitem[Alexanderson et~al.(2023)Alexanderson, Nagy, Beskow, and
  Henter]{alexanderson2023listen}
Simon Alexanderson, Rajmund Nagy, Jonas Beskow, and Gustav~Eje Henter.
\newblock Listen, denoise, action! audio-driven motion synthesis with diffusion
  models.
\newblock \emph{ACM Transactions on Graphics (TOG)}, 42\penalty0 (4):\penalty0
  1--20, 2023.

\bibitem[Ao et~al.(2022)Ao, Gao, Lou, Chen, and Liu]{ao2022rhythmic}
Tenglong Ao, Qingzhe Gao, Yuke Lou, Baoquan Chen, and Libin Liu.
\newblock Rhythmic gesticulator: Rhythm-aware co-speech gesture synthesis with
  hierarchical neural embeddings.
\newblock \emph{ACM Transactions on Graphics (TOG)}, 41\penalty0 (6):\penalty0
  1--19, 2022.

\bibitem[Ao et~al.(2023)Ao, Zhang, and Liu]{ao2023gesturediffuclip}
Tenglong Ao, Zeyi Zhang, and Libin Liu.
\newblock Gesturediffuclip: Gesture diffusion model with clip latents.
\newblock \emph{ACM Transactions on Graphics (TOG)}, 42\penalty0 (4):\penalty0
  1--18, 2023.

\bibitem[Bengio et~al.(2015)Bengio, Vinyals, Jaitly, and
  Shazeer]{bengio2015scheduled}
Samy Bengio, Oriol Vinyals, Navdeep Jaitly, and Noam Shazeer.
\newblock Scheduled sampling for sequence prediction with recurrent neural
  networks.
\newblock \emph{Advances in neural information processing systems}, 28, 2015.

\bibitem[Chen et~al.(2024{\natexlab{a}})Chen, Mart{\'\i}~Mons{\'o}, Du,
  Simchowitz, Tedrake, and Sitzmann]{chen2024diffusionforcing}
Boyuan Chen, Diego Mart{\'\i}~Mons{\'o}, Yilun Du, Max Simchowitz, Russ
  Tedrake, and Vincent Sitzmann.
\newblock Diffusion forcing: Next-token prediction meets full-sequence
  diffusion.
\newblock \emph{Advances in Neural Information Processing Systems},
  37:\penalty0 24081--24125, 2024{\natexlab{a}}.

\bibitem[Chen et~al.(2024{\natexlab{b}})Chen, Liu, Wang, Zeng, Li, and
  Chen]{chen2024diffsheg}
Junming Chen, Yunfei Liu, Jianan Wang, Ailing Zeng, Yu Li, and Qifeng Chen.
\newblock Diffsheg: A diffusion-based approach for real-time speech-driven
  holistic 3d expression and gesture generation.
\newblock In \emph{2024 IEEE/CVF Conference on Computer Vision and Pattern
  Recognition (CVPR)}, pages 7352--7361. IEEE, 2024{\natexlab{b}}.

\bibitem[Chen et~al.(2024{\natexlab{c}})Chen, Shi, Huang, Tan, Komura, and
  Chen]{chen2024camdm}
Rui Chen, Mingyi Shi, Shaoli Huang, Ping Tan, Taku Komura, and Xuelin Chen.
\newblock Taming diffusion probabilistic models for character control.
\newblock In \emph{ACM SIGGRAPH 2024 Conference Papers}, pages 1--10,
  2024{\natexlab{c}}.

\bibitem[Chhatre et~al.(2024)Chhatre, Athanasiou, Becherini, Peters, Black,
  Bolkart, et~al.]{chhatre2024amuse}
Kiran Chhatre, Nikos Athanasiou, Giorgio Becherini, Christopher Peters,
  Michael~J Black, Timo Bolkart, et~al.
\newblock Emotional speech-driven 3d body animation via disentangled latent
  diffusion.
\newblock In \emph{Proceedings of the IEEE/CVF conference on computer vision
  and pattern recognition}, pages 1942--1953, 2024.

\bibitem[D{\'e}fossez et~al.(2024)D{\'e}fossez, Mazar{\'e}, Orsini, Royer,
  P{\'e}rez, J{\'e}gou, Grave, and Zeghidour]{defossez2024moshi}
Alexandre D{\'e}fossez, Laurent Mazar{\'e}, Manu Orsini, Am{\'e}lie Royer,
  Patrick P{\'e}rez, Herv{\'e} J{\'e}gou, Edouard Grave, and Neil Zeghidour.
\newblock Moshi: a speech-text foundation model for real-time dialogue.
\newblock \emph{arXiv preprint arXiv:2410.00037}, 2024.

\bibitem[Ekstedt and Skantze(2020)]{ekstedt2020turngpt}
Erik Ekstedt and Gabriel Skantze.
\newblock Turngpt: a transformer-based language model for predicting
  turn-taking in spoken dialog.
\newblock In \emph{Findings of the Association for Computational Linguistics:
  EMNLP 2020}, pages 2981--2990, 2020.

\bibitem[Ekstedt and Skantze(2022)]{ekstedt2022vap}
Erik Ekstedt and Gabriel Skantze.
\newblock Voice activity projection: Self-supervised learning of turn-taking
  events.
\newblock \emph{arXiv preprint arXiv:2205.09812}, 2022.

\bibitem[Ferstl and McDonnell(2018)]{ferstl2018trinity}
Ylva Ferstl and Rachel McDonnell.
\newblock Investigating the use of recurrent motion modelling for speech
  gesture generation.
\newblock In \emph{Proceedings of the 18th international conference on
  intelligent virtual agents}, pages 93--98, 2018.

\bibitem[Fu et~al.(2026)Fu, Yu, Zhu, Luo, Chen, Shen, Song, Zhao, He, Hu, Shen,
  Ma, Huang, Liu, Zhang, and Chen]{fu2026cosa05}
Zhen Fu, Tao Yu, Hongbo Zhu, Haoxiang Luo, Zhiming Chen, Pinxi Shen, Jiaqi
  Song, Zheyi Zhao, Bozhen He, Jiangtao Hu, Cong Shen, Haolin Ma, Zimo Huang,
  Ben Liu, Wei Zhang, and Hua Chen.
\newblock Introducing {COSA} 0.5: A whole-body capability upgrade for the
  humanoid {VLA} {V3-0}.
\newblock \url{https://limxdynamics.com/cosa05v3/\#en}, 2026.
\newblock Published 2026-07-15.

\bibitem[Ghorbani et~al.(2023)Ghorbani, Ferstl, Holden, Troje, and
  Carbonneau]{ghorbani2023zeroeggs}
Saeed Ghorbani, Ylva Ferstl, Daniel Holden, Nikolaus~F Troje, and
  Marc-Andr{\'e} Carbonneau.
\newblock Zeroeggs: Zero-shot example-based gesture generation from speech.
\newblock In \emph{Computer Graphics Forum}, pages 206--216. Wiley Online
  Library, 2023.

\bibitem[Ginosar et~al.(2019)Ginosar, Bar, Kohavi, Chan, Owens, and
  Malik]{ginosar2019styles}
Shiry Ginosar, Amir Bar, Gefen Kohavi, Caroline Chan, Andrew Owens, and
  Jitendra Malik.
\newblock Learning individual styles of conversational gesture.
\newblock In \emph{2019 IEEE/CVF Conference on Computer Vision and Pattern
  Recognition (CVPR)}, pages 3492--3501. IEEE, 2019.

\bibitem[Guo et~al.(2024)Guo, Mu, Javed, Wang, and Cheng]{guo2024momask}
Chuan Guo, Yuxuan Mu, Muhammad~Gohar Javed, Sen Wang, and Li Cheng.
\newblock Momask: Generative masked modeling of 3d human motions.
\newblock In \emph{2024 IEEE/CVF Conference on Computer Vision and Pattern
  Recognition (CVPR)}, pages 1900--1910. IEEE, 2024.

\bibitem[Habibie et~al.(2021)Habibie, Xu, Mehta, Liu, Seidel, Pons-Moll,
  Elgharib, and Theobalt]{habibie2021video}
Ikhsanul Habibie, Weipeng Xu, Dushyant Mehta, Lingjie Liu, Hans-Peter Seidel,
  Gerard Pons-Moll, Mohamed Elgharib, and Christian Theobalt.
\newblock Learning speech-driven 3d conversational gestures from video.
\newblock In \emph{Proceedings of the 21st ACM international conference on
  intelligent virtual agents}, pages 101--108, 2021.

\bibitem[Ho and Salimans(2022)]{ho2022classifier}
Jonathan Ho and Tim Salimans.
\newblock Classifier-free diffusion guidance.
\newblock \emph{arXiv preprint arXiv:2207.12598}, 2022.

\bibitem[Huang et~al.(2026)Huang, Li, He, Zhou, and
  Shechtman]{huang2025selfforcing}
Xun Huang, Zhengqi Li, Guande He, Mingyuan Zhou, and Eli Shechtman.
\newblock Self forcing: Bridging the train-test gap in autoregressive video
  diffusion.
\newblock \emph{Advances in Neural Information Processing Systems},
  38:\penalty0 167283--167308, 2026.

\bibitem[Jang et~al.(2016)Jang, Gu, and Poole]{jang2016gumbel}
Eric Jang, Shixiang Gu, and Ben Poole.
\newblock Categorical reparameterization with gumbel-softmax.
\newblock \emph{arXiv preprint arXiv:1611.01144}, 2016.

\bibitem[Jiang et~al.(2023)Jiang, Chen, Liu, Yu, Yu, and
  Chen]{jiang2023motiongpt}
Biao Jiang, Xin Chen, Wen Liu, Jingyi Yu, Gang Yu, and Tao Chen.
\newblock Motiongpt: Human motion as a foreign language.
\newblock \emph{Advances in Neural Information Processing Systems},
  36:\penalty0 20067--20079, 2023.

\bibitem[Jiang et~al.(2025)Jiang, Xiao, Lin, Zhang, Ren, Gao, Lin, Cai, Yang,
  and Liu]{jiang2025solami}
Jianping Jiang, Weiye Xiao, Zhengyu Lin, Huaizhong Zhang, Tianxiang Ren, Yang
  Gao, Zhiqian Lin, Zhongang Cai, Lei Yang, and Ziwei Liu.
\newblock Solami: Social vision-language-action modeling for immersive
  interaction with 3d autonomous characters.
\newblock In \emph{2025 IEEE/CVF Conference on Computer Vision and Pattern
  Recognition (CVPR)}, pages 26887--26898. IEEE, 2025.

\bibitem[Kucherenko et~al.(2020)Kucherenko, Jonell, Van~Waveren, Henter,
  Alexandersson, Leite, and Kjellstr{\"o}m]{kucherenko2020gesticulator}
Taras Kucherenko, Patrik Jonell, Sanne Van~Waveren, Gustav~Eje Henter, Simon
  Alexandersson, Iolanda Leite, and Hedvig Kjellstr{\"o}m.
\newblock Gesticulator: A framework for semantically-aware speech-driven
  gesture generation.
\newblock In \emph{Proceedings of the 2020 international conference on
  multimodal interaction}, pages 242--250, 2020.

\bibitem[Lamb et~al.(2016)Lamb, Alias Parth~Goyal, Zhang, Zhang, Courville, and
  Bengio]{lamb2016professor}
Alex~M Lamb, Anirudh~Goyal Alias Parth~Goyal, Ying Zhang, Saizheng Zhang,
  Aaron~C Courville, and Yoshua Bengio.
\newblock Professor forcing: A new algorithm for training recurrent networks.
\newblock \emph{Advances in neural information processing systems}, 29, 2016.

\bibitem[Lee et~al.(2022)Lee, Kim, Kim, Cho, and Han]{lee2022rq}
Doyup Lee, Chiheon Kim, Saehoon Kim, Minsu Cho, and Wook-Shin Han.
\newblock Autoregressive image generation using residual quantization.
\newblock In \emph{2022 IEEE/CVF conference on computer vision and pattern
  recognition (CVPR)}, pages 11513--11522. IEEE, 2022.

\bibitem[Li et~al.(2021)Li, Kang, Pei, Zhe, Zhang, He, and
  Bao]{li2021audio2gestures}
Jing Li, Di Kang, Wenjie Pei, Xuefei Zhe, Ying Zhang, Zhenyu He, and Linchao
  Bao.
\newblock Audio2gestures: Generating diverse gestures from speech audio with
  conditional variational autoencoders.
\newblock In \emph{2021 IEEE/CVF International Conference on Computer Vision
  (ICCV)}, pages 11273--11282. IEEE, 2021.

\bibitem[Li et~al.(2017)Li, Zhou, Xiao, He, Huang, and Li]{li2017acrnn}
Zimo Li, Yi Zhou, Shuangjiu Xiao, Chong He, Zeng Huang, and Hao Li.
\newblock Auto-conditioned recurrent networks for extended complex human motion
  synthesis.
\newblock \emph{arXiv preprint arXiv:1707.05363}, 2017.

\bibitem[Liang et~al.(2024)Liang, Zhang, Li, Yu, and Xu]{liang2024intergen}
Han Liang, Wenqian Zhang, Wenxuan Li, Jingyi Yu, and Lan Xu.
\newblock Intergen: Diffusion-based multi-human motion generation under complex
  interactions.
\newblock \emph{International Journal of Computer Vision}, 132\penalty0
  (9):\penalty0 3463--3483, 2024.

\bibitem[{LimX Dynamics}(2025)]{limx_oli}
{LimX Dynamics}.
\newblock {LimX Oli}: Full-size general-purpose humanoid robot.
\newblock \url{https://www.limxdynamics.com/en/products/oli}, 2025.
\newblock Accessed: 2026-05-22.

\bibitem[{LimX Dynamics}(2026)]{limx_luna}
{LimX Dynamics}.
\newblock {LimX Luna}: Full-size interactive humanoid robot.
\newblock \url{https://www.limxdynamics.com/en/products/luna}, 2026.
\newblock Accessed: 2026-09-11.

\bibitem[Liu et~al.(2022{\natexlab{a}})Liu, Zhu, Iwamoto, Peng, Li, Zhou,
  Bozkurt, and Zheng]{liu2022beat}
Haiyang Liu, Zihao Zhu, Naoya Iwamoto, Yichen Peng, Zhengqing Li, You Zhou,
  Elif Bozkurt, and Bo Zheng.
\newblock Beat: A large-scale semantic and emotional multi-modal dataset for
  conversational gestures synthesis.
\newblock In \emph{European conference on computer vision}, pages 612--630.
  Springer, 2022{\natexlab{a}}.

\bibitem[Liu et~al.(2024{\natexlab{a}})Liu, Zhu, Becherini, Peng, Su, Zhou,
  Zhe, Iwamoto, Zheng, and Black]{liu2024emage}
Haiyang Liu, Zihao Zhu, Giorgio Becherini, Yichen Peng, Mingyang Su, You Zhou,
  Xuefei Zhe, Naoya Iwamoto, Bo Zheng, and Michael~J Black.
\newblock Emage: Towards unified holistic co-speech gesture generation via
  expressive masked audio gesture modeling.
\newblock In \emph{2024 IEEE/CVF Conference on Computer Vision and Pattern
  Recognition (CVPR)}, pages 1144--1154. IEEE, 2024{\natexlab{a}}.

\bibitem[Liu et~al.(2025)Liu, Song, Huang, Liu, and Xu]{liu2025gesturelsm}
Pinxin Liu, Luchuan Song, Junhua Huang, Haiyang Liu, and Chenliang Xu.
\newblock Gesturelsm: Latent shortcut based co-speech gesture generation with
  spatial-temporal modeling.
\newblock In \emph{2025 IEEE/CVF International Conference on Computer Vision
  (ICCV)}, pages 10929--10939. IEEE, 2025.

\bibitem[Liu et~al.(2026)Liu, Liu, Luo, Huang, Zou, and
  Song]{liu2026gesturefar}
Pinxin Liu, Haiyang Liu, Jiahao Luo, Junhua Huang, Chunhao Zou, and Luchuan
  Song.
\newblock Gesturefar: Streaming co-speech gesture generation with flow
  autoregression.
\newblock \emph{arXiv preprint arXiv:2609.21576}, 2026.

\bibitem[Liu et~al.(2022{\natexlab{b}})Liu, Wu, Zhou, Xu, Qian, Lin, Zhou, Wu,
  Dai, and Zhou]{liu2022ha2g}
Xian Liu, Qianyi Wu, Hang Zhou, Yinghao Xu, Rui Qian, Xinyi Lin, Xiaowei Zhou,
  Wayne Wu, Bo Dai, and Bolei Zhou.
\newblock Learning hierarchical cross-modal association for co-speech gesture
  generation.
\newblock In \emph{2022 IEEE/CVF Conference on Computer Vision and Pattern
  Recognition (CVPR)}, pages 10452--10462. IEEE, 2022{\natexlab{b}}.

\bibitem[Liu et~al.(2024{\natexlab{b}})Liu, Cao, Wen, Jiang, and
  Ding]{liu2024probtalk}
Yifei Liu, Qiong Cao, Yandong Wen, Huaiguang Jiang, and Changxing Ding.
\newblock Towards variable and coordinated holistic co-speech motion
  generation.
\newblock In \emph{2024 IEEE/CVF Conference on Computer Vision and Pattern
  Recognition (CVPR)}, pages 1566--1576. IEEE, 2024{\natexlab{b}}.

\bibitem[McLean et~al.(2025)McLean, Meendering, Swartz, Gabbay, Olsen, Jacobs,
  Rosen, de~Bree, Garcia, Merrill, et~al.]{mclean2025embody}
Claire McLean, Makenzie Meendering, Tristan Swartz, Orri Gabbay, Alexandra
  Olsen, Rachel Jacobs, Nicholas Rosen, Philippe de Bree, Tony Garcia, Gadsden
  Merrill, et~al.
\newblock Embody 3d: A large-scale multimodal motion and behavior dataset.
\newblock \emph{arXiv preprint arXiv:2510.16258}, 2025.

\bibitem[Mughal et~al.(2024)Mughal, Dabral, Habibie, Donatelli, Habermann, and
  Theobalt]{mughal2024convofusion}
Muhammad~Hamza Mughal, Rishabh Dabral, Ikhsanul Habibie, Lucia Donatelli, Marc
  Habermann, and Christian Theobalt.
\newblock Convofusion: Multi-modal conversational diffusion for co-speech
  gesture synthesis.
\newblock In \emph{2024 IEEE/CVF Conference on Computer Vision and Pattern
  Recognition (CVPR)}, pages 1388--1398. IEEE, 2024.

\bibitem[Mughal et~al.(2026)Mughal, Dabral, Demberg, and
  Theobalt]{mughal2026miburi}
M~Hamza Mughal, Rishabh Dabral, Vera Demberg, and Christian Theobalt.
\newblock Miburi: Towards expressive interactive gesture synthesis.
\newblock \emph{arXiv preprint arXiv:2603.03282}, 2026.

\bibitem[Nagano et~al.(2026)Nagano, Liu, Park, Li, Mazumdar, Jacobsen, Wang,
  Stengel, Roy, Cheung, et~al.]{nagano2026dyaplex}
Koki Nagano, Hongyu Liu, Seonwook Park, Tianye Li, Amrita Mazumdar, Christian
  Jacobsen, Shengze Wang, Michael Stengel, Rajarshi Roy, Ka~Chun Cheung, et~al.
\newblock Dyaplex: Full-duplex speech-motion model for dyadic interaction.
\newblock \emph{arXiv preprint arXiv:2606.03874}, 2026.

\bibitem[Ng et~al.(2022)Ng, Joo, Hu, Li, Darrell, Kanazawa, and
  Ginosar]{ng2022learning}
Evonne Ng, Hanbyul Joo, Liwen Hu, Hao Li, Trevor Darrell, Angjoo Kanazawa, and
  Shiry Ginosar.
\newblock Learning to listen: Modeling non-deterministic dyadic facial motion.
\newblock In \emph{2022 IEEE/CVF Conference on Computer Vision and Pattern
  Recognition (CVPR)}, pages 20363--20373. IEEE, 2022.

\bibitem[Ng et~al.(2024)Ng, Romero, Bagautdinov, Bai, Darrell, Kanazawa, and
  Richard]{ng2024audio2photoreal}
Evonne Ng, Javier Romero, Timur Bagautdinov, Shaojie Bai, Trevor Darrell,
  Angjoo Kanazawa, and Alexander Richard.
\newblock From audio to photoreal embodiment: Synthesizing humans in
  conversations.
\newblock In \emph{2024 IEEE/CVF Conference on Computer Vision and Pattern
  Recognition (CVPR)}, pages 1001--1010. IEEE, 2024.

\bibitem[Nguyen et~al.(2023)Nguyen, Kharitonov, Copet, Adi, Hsu, Elkahky,
  Tomasello, Algayres, Sagot, Mohamed, et~al.]{nguyen2023dgslm}
Tu~Anh Nguyen, Eugene Kharitonov, Jade Copet, Yossi Adi, Wei-Ning Hsu, Ali
  Elkahky, Paden Tomasello, Robin Algayres, Benoit Sagot, Abdelrahman Mohamed,
  et~al.
\newblock Generative spoken dialogue language modeling.
\newblock \emph{Transactions of the Association for Computational Linguistics},
  11:\penalty0 250--266, 2023.

\bibitem[Pavlakos et~al.(2019)Pavlakos, Choutas, Ghorbani, Bolkart, Osman,
  Tzionas, and Black]{pavlakos2019smplx}
Georgios Pavlakos, Vasileios Choutas, Nima Ghorbani, Timo Bolkart, Ahmed~AA
  Osman, Dimitrios Tzionas, and Michael~J Black.
\newblock Expressive body capture: 3d hands, face, and body from a single
  image.
\newblock In \emph{Proceedings of the IEEE/CVF conference on computer vision
  and pattern recognition}, pages 10975--10985, 2019.

\bibitem[Peng et~al.(2026)Peng, Song, Jung, Liu, Liu, Chu, Liu, Wu, Koike, and
  Kitani]{peng2026dyadit}
Yichen Peng, Jyun-Ting Song, Siyeol Jung, Ruofan Liu, Haiyang Liu, Xuangeng
  Chu, Ruicong Liu, Erwin Wu, Hideki Koike, and Kris Kitani.
\newblock Dyadit: A multi-modal diffusion transformer for socially favorable
  dyadic gesture generation.
\newblock \emph{arXiv preprint arXiv:2602.23165}, 2026.

\bibitem[Peng et~al.(2025)Peng, Fan, Wu, Wang, Liu, He, and
  Fan]{peng2025dualtalk}
Ziqiao Peng, Yanbo Fan, Haoyu Wu, Xuan Wang, Hongyan Liu, Jun He, and Zhaoxin
  Fan.
\newblock Dualtalk: Dual-speaker interaction for 3d talking head conversations.
\newblock In \emph{2025 IEEE/CVF Conference on Computer Vision and Pattern
  Recognition (CVPR)}, pages 21055--21064. IEEE, 2025.

\bibitem[Perez et~al.(2018)Perez, Strub, De~Vries, Dumoulin, and
  Courville]{perez2018film}
Ethan Perez, Florian Strub, Harm De~Vries, Vincent Dumoulin, and Aaron
  Courville.
\newblock Film: Visual reasoning with a general conditioning layer.
\newblock In \emph{Proceedings of the AAAI conference on artificial
  intelligence}, 2018.

\bibitem[Qian et~al.(2021)Qian, Tu, Zhi, Liu, and Gao]{qian2021templates}
Shenhan Qian, Zhi Tu, Yihao Zhi, Wen Liu, and Shenghua Gao.
\newblock Speech drives templates: Co-speech gesture synthesis with learned
  templates.
\newblock In \emph{2021 IEEE/CVF International Conference on Computer Vision
  (ICCV)}, pages 11057--11066. IEEE, 2021.

\bibitem[Ranzato et~al.(2015)Ranzato, Chopra, Auli, and
  Zaremba]{ranzato2016sequence}
Marc'Aurelio Ranzato, Sumit Chopra, Michael Auli, and Wojciech Zaremba.
\newblock Sequence level training with recurrent neural networks.
\newblock \emph{arXiv preprint arXiv:1511.06732}, 2015.

\bibitem[Ross et~al.(2011)Ross, Gordon, and Bagnell]{ross2011dagger}
St{\'e}phane Ross, Geoffrey Gordon, and Drew Bagnell.
\newblock A reduction of imitation learning and structured prediction to
  no-regret online learning.
\newblock In \emph{Proceedings of the fourteenth international conference on
  artificial intelligence and statistics}, pages 627--635. JMLR Workshop and
  Conference Proceedings, 2011.

\bibitem[Saleem et~al.(2026)Saleem, Patel, Pinyoanuntapong, Qin, Yang, Xue,
  Helmy, Chen, and Wang]{saleem2026livegesture}
Muhammad~Usama Saleem, Mayur~Jagdishbhai Patel, Ekkasit Pinyoanuntapong,
  Zhongxing Qin, Li Yang, Hongfei Xue, Ahmed Helmy, Chen Chen, and Pu Wang.
\newblock Livegesture streamable co-speech gesture generation model.
\newblock \emph{arXiv preprint arXiv:2604.10927}, 2026.

\bibitem[Shi et~al.(2024{\natexlab{a}})Shi, Qin, Ho, Liao, Huang, Yamagishi,
  and Komura]{shi2024ittakestwo}
Mingyi Shi, Dafei Qin, Leo Ho, Zhouyingcheng Liao, Yinghao Huang, Junichi
  Yamagishi, and Taku Komura.
\newblock It takes two: Real-time co-speech two-person's interaction generation
  via reactive auto-regressive diffusion model.
\newblock \emph{arXiv preprint arXiv:2412.02419}, 2024{\natexlab{a}}.

\bibitem[Shi et~al.(2024{\natexlab{b}})Shi, Wang, Jiang, Lin, Dai, and
  Peng]{shi2024amdm}
Yi Shi, Jingbo Wang, Xuekun Jiang, Bingkun Lin, Bo Dai, and Xue~Bin Peng.
\newblock Interactive character control with auto-regressive motion diffusion
  models.
\newblock \emph{ACM Transactions On Graphics (TOG)}, 43\penalty0 (4):\penalty0
  1--14, 2024{\natexlab{b}}.

\bibitem[Su et~al.(2024)Su, Ahmed, Lu, Pan, Bo, and Liu]{su2021roformer}
Jianlin Su, Murtadha Ahmed, Yu Lu, Shengfeng Pan, Wen Bo, and Yunfeng Liu.
\newblock Roformer: Enhanced transformer with rotary position embedding.
\newblock \emph{Neurocomputing}, 568:\penalty0 127063, 2024.

\bibitem[Sun et~al.(2025)Sun, Xu, Jiang, Liu, Sun, and Huang]{sun2024beyond}
Mingze Sun, Chao Xu, Xinyu Jiang, Yang Liu, Baigui Sun, and Ruqi Huang.
\newblock Beyond talking--generating holistic 3d human dyadic motion for
  communication.
\newblock \emph{International Journal of Computer Vision}, 133\penalty0
  (5):\penalty0 2910--2926, 2025.

\bibitem[Tevet et~al.(2022)Tevet, Raab, Gordon, Shafir, Cohen-Or, and
  Bermano]{tevet2023mdm}
Guy Tevet, Sigal Raab, Brian Gordon, Yonatan Shafir, Daniel Cohen-Or, and
  Amit~H Bermano.
\newblock Human motion diffusion model.
\newblock \emph{arXiv preprint arXiv:2209.14916}, 2022.

\bibitem[Tevet et~al.(2025)Tevet, Raab, Cohan, Reda, Luo, Peng, Bermano, and
  Van~de Panne]{tevet2025closd}
Guy Tevet, Sigal Raab, Setareh Cohan, Daniele Reda, Zhengyi Luo, Xue~Bin Peng,
  Amit Bermano, and Michiel Van~de Panne.
\newblock Closd: Closing the loop between simulation and diffusion for
  multi-task character control.
\newblock In \emph{International Conference on Learning Representations}, pages
  46506--46520, 2025.

\bibitem[Tran et~al.(2024)Tran, Chang, Siniukov, and Soleymani]{tran2024dim}
Minh Tran, Di Chang, Maksim Siniukov, and Mohammad Soleymani.
\newblock Dim: Dyadic interaction modeling for social behavior generation.
\newblock In \emph{European Conference on Computer Vision}, pages 484--503.
  Springer, 2024.

\bibitem[Van Den~Oord et~al.(2017)Van Den~Oord, Vinyals, et~al.]{oord2017vqvae}
Aaron Van Den~Oord, Oriol Vinyals, et~al.
\newblock Neural discrete representation learning.
\newblock \emph{Advances in neural information processing systems}, 30, 2017.

\bibitem[Veluri et~al.(2024)Veluri, Peloquin, Yu, Gong, and
  Gollakota]{veluri2024syncllm}
Bandhav Veluri, Benjamin~N Peloquin, Bokai Yu, Hongyu Gong, and Shyamnath
  Gollakota.
\newblock Beyond turn-based interfaces: Synchronous llms as full-duplex
  dialogue agents.
\newblock In \emph{Proceedings of the 2024 Conference on Empirical Methods in
  Natural Language Processing}, pages 21390--21402, 2024.

\bibitem[Wang et~al.(2026{\natexlab{a}})Wang, Hu, Yu, and Yu]{wang2026timo}
Zhao Wang, Jiangtao Hu, Jack Yu, and Tao Yu.
\newblock Timo: Taming multimodal diffusion transformers for human motion
  generation.
\newblock \emph{arXiv preprint arXiv:2609.30761}, 2026{\natexlab{a}}.

\bibitem[Wang et~al.(2026{\natexlab{b}})Wang, Ren, Shi, Wang, Lin, Wang, Qi,
  Zhao, Wang, and Yi]{wang2026robogesture}
Zifan Wang, Ziang Ren, Pengyang Shi, Zirui Wang, Chenghuai Lin, Tianze Wang,
  Zekun Qi, Liangliang Zhao, He Wang, and Li Yi.
\newblock Robogesture: Real-time semantic-aligned co-speech gestures generation
  for humanoid interaction.
\newblock In \emph{European Conference on Computer Vision}, pages 112--130.
  Springer, 2026{\natexlab{b}}.

\bibitem[Xiao et~al.(2025)Xiao, Lu, Pi, Fan, Pan, Zhou, Feng, Zhou, Peng, and
  Wang]{xiao2025motionstreamer}
Lixing Xiao, Shunlin Lu, Huaijin Pi, Ke Fan, Liang Pan, Yueer Zhou, Ziyong
  Feng, Xiaowei Zhou, Sida Peng, and Jingbo Wang.
\newblock Motionstreamer: Streaming motion generation via diffusion-based
  autoregressive model in causal latent space.
\newblock In \emph{2025 IEEE/CVF International Conference on Computer Vision
  (ICCV)}, pages 10086--10096. IEEE, 2025.

\bibitem[Yang et~al.(2023{\natexlab{a}})Yang, Wu, Li, Zhang, Hao, Bao, Cheng,
  and Xiao]{yang2023diffusestylegesture}
Sicheng Yang, Zhiyong Wu, Minglei Li, Zhensong Zhang, Lei Hao, Weihong Bao,
  Ming Cheng, and Long Xiao.
\newblock Diffusestylegesture: Stylized audio-driven co-speech gesture
  generation with diffusion models.
\newblock \emph{arXiv preprint arXiv:2305.04919}, 2023{\natexlab{a}}.

\bibitem[Yang et~al.(2023{\natexlab{b}})Yang, Wu, Li, Zhang, Hao, Bao, and
  Zhuang]{yang2023qpgesture}
Sicheng Yang, Zhiyong Wu, Minglei Li, Zhensong Zhang, Lei Hao, Weihong Bao, and
  Haolin Zhuang.
\newblock Qpgesture: Quantization-based and phase-guided motion matching for
  natural speech-driven gesture generation.
\newblock In \emph{2023 IEEE/CVF Conference on Computer Vision and Pattern
  Recognition (CVPR)}, pages 2321--2330. IEEE, 2023{\natexlab{b}}.

\bibitem[Yi et~al.(2023)Yi, Liang, Liu, Cao, Wen, Bolkart, Tao, and
  Black]{yi2023talkshow}
Hongwei Yi, Hualin Liang, Yifei Liu, Qiong Cao, Yandong Wen, Timo Bolkart,
  Dacheng Tao, and Michael~J Black.
\newblock Generating holistic 3d human motion from speech.
\newblock In \emph{2023 IEEE/CVF Conference on Computer Vision and Pattern
  Recognition (CVPR)}, pages 469--480. IEEE, 2023.

\bibitem[Yoon et~al.(2020)Yoon, Cha, Lee, Jang, Lee, Kim, and
  Lee]{yoon2020trimodal}
Youngwoo Yoon, Bok Cha, Joo-Haeng Lee, Minsu Jang, Jaeyeon Lee, Jaehong Kim,
  and Geehyuk Lee.
\newblock Speech gesture generation from the trimodal context of text, audio,
  and speaker identity.
\newblock \emph{ACM Transactions on Graphics (TOG)}, 39\penalty0 (6):\penalty0
  1--16, 2020.

\bibitem[Zhang et~al.(2023)Zhang, Zhang, Cun, Zhang, Zhao, Lu, Shen, and
  Shan]{zhang2023t2mgpt}
Jianrong Zhang, Yangsong Zhang, Xiaodong Cun, Yong Zhang, Hongwei Zhao, Hongtao
  Lu, Xi Shen, and Ying Shan.
\newblock Generating human motion from textual descriptions with discrete
  representations.
\newblock In \emph{Proceedings of the IEEE/CVF conference on computer vision
  and pattern recognition}, pages 14730--14740, 2023.

\bibitem[Zhang et~al.(2025)Zhang, Li, Zhang, Dang, Ren, Bo, and
  Tu]{zhang2025semtalk}
Xiangyue Zhang, Jianfang Li, Jiaxu Zhang, Ziqiang Dang, Jianqiang Ren, Liefeng
  Bo, and Zhigang Tu.
\newblock Semtalk: Holistic co-speech motion generation with frame-level
  semantic emphasis.
\newblock In \emph{2025 IEEE/CVF International Conference on Computer Vision
  (ICCV)}, pages 13761--13771. IEEE, 2025.

\bibitem[Zhang et~al.(2024)Zhang, Ao, Zhang, Gao, Lin, Chen, and
  Liu]{zhang2024semgesticulator}
Zeyi Zhang, Tenglong Ao, Yuyao Zhang, Qingzhe Gao, Chuan Lin, Baoquan Chen, and
  Libin Liu.
\newblock Semantic gesticulator: Semantics-aware co-speech gesture synthesis.
\newblock \emph{ACM Transactions on Graphics (TOG)}, 43\penalty0 (4):\penalty0
  1--17, 2024.

\bibitem[Zhao et~al.(2025)Zhao, Li, and Tang]{zhao2025dart}
Kaifeng Zhao, Gen Li, and Siyu Tang.
\newblock Dartcontrol: A diffusion-based autoregressive motion model for
  real-time text-driven motion control.
\newblock In \emph{International Conference on Learning Representations}, pages
  23569--23592, 2025.

\bibitem[Zhao et~al.(2026)Zhao, Xiao, and Chen]{zhao2026kinest}
Shuting Zhao, Zeyu Xiao, and Xinrong Chen.
\newblock Kinest: A kinematics-guided spatiotemporal state space model for
  human motion tracking from sparse signals.
\newblock In \emph{Proceedings of the AAAI Conference on Artificial
  Intelligence}, pages 13244--13252, 2026.

\bibitem[Zhi et~al.(2023)Zhi, Cun, Chen, Shen, Guo, Huang, and
  Gao]{zhi2023livelyspeaker}
Yihao Zhi, Xiaodong Cun, Xuelin Chen, Xi Shen, Wen Guo, Shaoli Huang, and
  Shenghua Gao.
\newblock Livelyspeaker: Towards semantic-aware co-speech gesture generation.
\newblock In \emph{2023 IEEE/CVF International Conference on Computer Vision
  (ICCV)}, pages 20750--20760. IEEE, 2023.

\bibitem[Zhou et~al.(2022)Zhou, Bai, Zhang, Yao, Zhao, and Mei]{zhou2022vico}
Mohan Zhou, Yalong Bai, Wei Zhang, Ting Yao, Tiejun Zhao, and Tao Mei.
\newblock Responsive listening head generation: a benchmark dataset and
  baseline.
\newblock In \emph{European conference on computer vision}, pages 124--142.
  Springer, 2022.

\bibitem[Zhou et~al.(2019)Zhou, Barnes, Lu, Yang, and Li]{zhou2019continuity}
Yi Zhou, Connelly Barnes, Jingwan Lu, Jimei Yang, and Hao Li.
\newblock On the continuity of rotation representations in neural networks.
\newblock In \emph{2019 IEEE/CVF Conference on Computer Vision and Pattern
  Recognition (CVPR)}, pages 5738--5746. IEEE, 2019.

\bibitem[Zhu et~al.(2023)Zhu, Liu, Liu, Qian, Liu, and Yu]{zhu2023diffgesture}
Lingting Zhu, Xian Liu, Xuanyu Liu, Rui Qian, Ziwei Liu, and Lequan Yu.
\newblock Taming diffusion models for audio-driven co-speech gesture
  generation.
\newblock In \emph{2023 IEEE/CVF Conference on Computer Vision and Pattern
  Recognition (CVPR)}, pages 10544--10553. IEEE, 2023.

\bibitem[Zhu et~al.(2025)Zhu, Zhang, Rong, Hu, Liang, and Ge]{zhu2025infp}
Yongming Zhu, Longhao Zhang, Zhengkun Rong, Tianshu Hu, Shuang Liang, and
  Zhipeng Ge.
\newblock Infp: Audio-driven interactive head generation in dyadic
  conversations.
\newblock In \emph{2025 IEEE/CVF Conference on Computer Vision and Pattern
  Recognition (CVPR)}, pages 10667--10677. IEEE, 2025.

\end{thebibliography}
}
\clearpage
\maketitlesupplementary
\appendix
\setcounter{dbltopnumber}{3}
\setcounter{topnumber}{3}
\setcounter{totalnumber}{5}
\renewcommand{\topfraction}{0.6}
\makeatletter
\setlength{\@fptop}{0pt}
\setlength{\@dblfptop}{0pt}
\setlength{\@dblfpsep}{16pt}
\setlength{\@dblfpbot}{0pt plus 1fil}
\makeatother

\section{Dataset Construction}
\label{sec:supp_data}
\paragraph{Corpus construction.}
Tab.~\ref{tab:data} lists the corpus composition, and Fig.~\ref{fig:teaser}(b) shows its speech activity.
All motion is converted to SMPL-X at 25\,fps.
Seamless Interaction provides most of the conversational data through monocular reconstruction, and since its finger and facial detail is less reliable than body pose~\cite{agrawal2025seamless,nagano2026dyaplex}, we exclude fingers and face. For the data filtering, we remove frames with implausible wrist flips or self-penetration of the forearms or hands into the torso, head, or the other arm.
Penetration is measured on the SMPL-X mesh, with thresholds of 30\,mm into the torso and 10\,mm between arms.
We also remove walking frames with root speed above 25\,cm/s or steps over 15\,cm.
During mutual silence, we remove frames with an upper arm raised above $45^\circ$ or wrist speed above 100\,cm/s.
Seated and multi-party Embody 3D recordings are excluded.
Only contiguous segments of at least 10\,s are retained.
Within each source, we split recordings into training and held-out partitions at approximately 85:15, keeping both participants and all segments of a recording in one partition.
The codec and generator use the 1{,}371\,h training set.

\paragraph{Conversation samples and activity labels.}
Each conversation in Seamless Interaction and Embody 3D yields one sample per participant whose motion passes filtering, with that participant's audio as the self stream and the other's as the partner stream.
Audio2Photoreal provides motion for one participant and separate audio channels for self and partner, and BEAT2 and Trinity provide monologue samples with a null partner stream.
Speech loudness is normalized per file to $-23$\,dBFS.
For Seamless Interaction, we use the released voice-activity annotations as reference labels.
For the other sources, we derive labels from 40\,ms frames of each unscaled speaker track.
Frames of Embody 3D and Audio2Photoreal are active above $-50$\,dBFS, and those of BEAT2 and Trinity above a threshold 30\,dB below the segment's 95th-percentile frame energy.

\begin{table}[t]
\centering
\caption{\textbf{Corpus composition.} Recording counts refer to conversations or monologues. Samples and hours are counted per retained target participant. The first three sources are conversational, and BEAT2 and Trinity contain monologues.}
\label{tab:data}
\small
\setlength{\tabcolsep}{3pt}
\resizebox{\linewidth}{!}{\begin{tabular}{lrrrrr}
\toprule
Source & Recordings & Samples & Train (h) & Held-out (h) & Total (h)\\
\midrule
Seamless Interaction & 43{,}100 & 62{,}427 & 1{,}339.7 & 236.3 & 1{,}576.0\\
Embody 3D & 468 & 704 & 8.0 & 1.9 & 9.9\\
Audio2Photoreal & 78 & 78 & 2.1 & 0.5 & 2.6\\
BEAT2 & 1{,}208 & 1{,}208 & 20.3 & 4.2 & 24.5\\
Trinity & 24 & 24 & 1.1 & 0.2 & 1.3\\
\midrule
Total & 44{,}878 & 64{,}441 & 1{,}371.2 & 243.1 & 1{,}614.3\\
\bottomrule
\end{tabular}}
\end{table}

\section{Implementation Details}
\label{sec:supp_impl}
\paragraph{Architecture.}
The causal motion codec follows the Mimi design~\cite{defossez2024moshi}.
Its encoder maps the 6D rotations of the 13 joints through a causal convolutional stem with two residual blocks and a stride-2 downsampling layer of width 256, followed by an 8-layer causal Transformer of width 256 with 4 heads and a feed-forward width of 2{,}048; the decoder mirrors the encoder.
The latent representation of each pair of frames is quantized by $K{=}8$ RVQ levels, each with 2{,}048 entries of dimension 128.
In the generator, each speech stream is given by the 8 Mimi codebooks at 12.5\,Hz, embedded with the 4{,}096-dimensional input embedding tables of Moshi~\cite{defossez2024moshi}, summed over codebooks, and projected to width 256 by the stream-specific projection $P_r$.
Motion codes are embedded with tables initialized from the codec codebooks and projected to the same width.
The temporal Transformer has 4 layers of width 256 with 2 heads, SiLU-gated feed-forward layers of width 672, query--key normalization, and dropout 0.1.
Speech slots are visible for $L_{\mathrm{s}}{=}L_{\mathrm{p}}{=}50$ steps (4\,s) and motion slots for $L_{\mathrm{m}}{=}25$ steps (2\,s) in Eq.~\ref{eq:stream_mask}, so each step attends to at most 125 slots; the key--value cache keeps all slots of the last 50 steps and masks older motion slots.
The turn-taking cues use the recency cap $d_{\max}{=}30$\,s and are mapped by a two-layer SiLU MLP of width 256, whose output layers for $\mathbf b_t$ and the feature-wise scale and shift are zero-initialized.
The depth Transformer has 2 layers of width 256 with one head and level-specific weights for its 7 levels, and attends to the speech embeddings of both streams at the current step.

\paragraph{Training.}
The codec is trained for 30 epochs on 8 GPUs with a batch size of 16 per GPU, using AdamW with a learning rate of $2{\times}10^{-4}$ and quantizer dropout.
Its loss combines reconstruction errors on 6D rotations, axis angles, and rotation matrices with errors on SMPL-X joint positions and mesh vertices and their temporal differences, plus the RVQ commitment loss.
The codec is then frozen, and the generator is trained with AdamW at a constant learning rate of $4{\times}10^{-4}$ and $\beta{=}(0.9,0.95)$, with batch size 64 per GPU on 8 NVIDIA L20Z GPUs in bf16, for 170 epochs, which takes about 17\,h.
Mixed-history training starts at epoch 15, and $\epsilon$ then ramps linearly to $\epsilon_{\max}{=}0.2$.
The turn-taking cues are computed over the whole recording before cropping, so each training crop starts with its true speaking history, as at inference.
Self audio, partner audio, the turn-taking cues, and the style embedding are dropped with probabilities 0.1, 0.15, 0.1, and 0.1.
For the objectives of Sec.~\ref{sec:kinematic}, we use a Gumbel temperature $\tau{=}1$, Smooth-L1 $\beta{=}0.05$, $\lambda_{\mathrm{dyn}}{=}30$, $\lambda_{\mathrm{reg}}{=}10$, and a $z$-loss weight of $10^{-4}$.
The interval sets $\mathcal F$ in Eq.~\ref{eq:dynamics_losses} comprise all valid frames, low-motion frames with a reference arm speed below $\delta{=}0.008$\,rad/frame, and silent frames with zero self and partner activity, with all weights $\lambda_f$ set to one.
The sets may overlap, and each is normalized separately, which keeps short low-motion and silent intervals from being outweighed by active motion.
Rotational-increment errors are averaged over consecutive frame pairs inside a set and temporal-difference errors over consecutive frame triples, both over all generated joints and rotation-vector components, and an empty set contributes zero.
The movement regularization of Eq.~\ref{eq:reg_losses} uses 25-frame windows, a speed margin $\gamma{=}1.05$, and weights $\lambda_{\mathrm{spd}}{=}1$ and $\lambda_{\mathrm{hold}}{=}3$, and reference holds use the threshold $\delta$ above.

\paragraph{Inference.}
At evaluation, the motion-style embedding $\mathbf e_{\mathrm{id}}$ is set to each clip's data source.
Classifier-free guidance with scale 4 contrasts the full condition with null self- and partner-audio tokens and keeps the turn-taking cues in both branches.
We sample with temperature 0.9 and top-$p$ 0.8, or 0.95 in the depth model.

\section{Evaluation Protocol}
\label{sec:supp_metrics}
This section describes the evaluation data, namely our evaluation set and the BEAT2 benchmark, and the raw metrics behind the five score dimensions.
All metrics are computed for generated and GT motion with the same forward kinematics on the 13 generated joints at 25\,fps.
Unless stated otherwise, a metric is scored in $[0,100]$ by $100\min(r,1/r)$ for its ratio $r$ to GT, with raw values in Sec.~\ref{sec:supp_raw}.

\paragraph{Evaluation set.}
We randomly sample 200 held-out clips of up to 90\,s.
These comprise 100 Seamless Interaction, 50 Embody 3D, 30 BEAT2, 10 Trinity, and 10 Audio2Photoreal clips.
Of the 160 clips from conversational sources, 136 contain partner speech and form the conversation subset.
In the other 24, only the target participant speaks, so they join the 40 BEAT2 and Trinity clips to form the 64-clip monologue subset.

\paragraph{BEAT2 benchmark.}
For Tab.~\ref{tab:beat2}, all methods are trained on the official BEAT2 training split and evaluated on its test split.
They generate from audio alone without GT seed frames and are scored on the same 13 upper-body joints.
FGD uses the official autoencoder with the other joints set to zero, so numbers differ from the original papers.

\paragraph{Rhythm.}
Beat consistency (BC) measures how closely motion beats follow self-speech onsets.
Only onsets on frames where the target participant speaks are used.
Onsets while listening or in silence come from partner bleed-through and noise.
Its score is $100\exp(-|\ln r|/\tau)$, with $\tau{=}0.75$ for $r{<}1$ and $\tau{=}1.5$ otherwise.

\paragraph{Naturalness.}
The Fr\'echet gesture distance (FGD) is computed in the latent space of an upper-body motion encoder trained on our data~\cite{peng2026dyadit}, and the score is $100\exp(-\mathrm{FGD}/5)$.
The raw tables also list Fr\'echet distances in joint-position (FD-S) and joint-velocity (FD-K) space, which do not enter the score.

\paragraph{Smoothness.}
Average jerk (AJ), peak jerk (PJ), and area under jerk (AUJ) are computed on joint rotations over each sequence and calibrated against GT.
Values below GT may reflect overly damped motion, and values above GT may reflect jitter.
Since PJ and AUJ are sensitive to a few extreme frames, the scores are weighted by 0.7, 0.15, and 0.15.

\paragraph{Diversity.}
Within-clip L1 variance~\cite{mughal2026miburi} is computed separately for the arms and for the torso and head, so that excess torso sway cannot hide missing arm motion.
The score averages the two group scores.

\paragraph{Turn-taking.}
Listening Motion Activity is the mean joint speed $v_t$ over listening frames, where the partner speaks and the target participant does not, $\mathrm{LMA}{=}\sum_t \ell_t v_t/\sum_t \ell_t$ with $\ell_t{=}1$ on these frames.
Interaction Timing Alignment (ITA) scores each partner speech end $e$ followed by at least 0.5\,s of partner silence by how well the BC motion beats $\mathcal M_j$ of the $J$ joints follow it,
\begin{equation}
g(e)=\frac{1}{J}\sum_{j=1}^{J}\max_{\substack{m\in\mathcal M_j\\0\le m-e\le 2.5\,\mathrm s}}\exp\Big(-\frac{(m-e-\delta)^2}{2\sigma^2}\Big),
\end{equation}
where a joint without such a beat contributes zero, and $\delta{=}1.05$\,s and $\sigma{=}0.74$\,s are the median and spread of GT beat lags.
To remove the effect of beat density, ITA averages $g(e)-c$ over events, where $c$ is $g$ averaged over all frames of the clip.
The score averages the LMA and ITA scores on the conversation clips.

\section{Compared Methods}
\label{sec:supp_baselines}
All methods are trained on the training partition of Tab.~\ref{tab:data} with the 13-joint upper-body representation and evaluated on the same 200 clips with the same metrics.
We follow the official training and inference recipes, including the GT seed frames with which SemTalk and GestureLSM start generation.
On our corpus, word inputs are removed since most sources lack word-aligned transcripts, and speaker identities are replaced by the data source.
On BEAT2, each method retains its original architecture and optimization settings. All methods use audio-only input without GT seed frames and follow the evaluation protocol in Sec.~\ref{sec:supp_metrics}.

\paragraph{DiffSHEG~\cite{chen2024diffsheg}.}
A diffusion Transformer that jointly generates expressions and gestures from speech and extends to long sequences by outpainting each chunk from the end of the previous one.
Since our motion excludes the face, we retrain its gesture generator on our data.

\paragraph{SemTalk~\cite{zhang2025semtalk}.}
A two-stage RVQ model that generates rhythm-related base motion from speech and adds sparse semantic gestures from word-aligned transcripts.
We retrain its speech-driven base stage on HuBERT features.

\paragraph{GestureLSM~\cite{liu2025gesturelsm}.}
A latent shortcut model that generates the latents of an RVQ-VAE in a few sampling steps.
We retrain both its RVQ-VAE and its shortcut model on our data.

\paragraph{MIBURI~\cite{mughal2026miburi}.}
A real-time model in which a frozen full-duplex speech model drives the autoregressive generation of RVQ motion codes.
We retrain both its motion codec and its gesture model on our data.

\paragraph{Audio2Photoreal~\cite{ng2024audio2photoreal}.}
A conversational model that generates one participant's body motion from both participants' audio, with a Transformer that predicts guide keyframes and a diffusion model that completes the motion.
We retrain its body model on both audio streams of our data.

\paragraph{DyaPlex~\cite{nagano2026dyaplex}.}
A dual-tower model in which a full-duplex speech model drives a causal motion Transformer that predicts the RVQ codes of both participants.
We reimplement it with Moshi~\cite{defossez2024moshi} as the speech model and, following its teacher-forced configuration, give the partner's GT motion as observed input, or a null motion for monologues.

\section{Additional Quantitative Results}
\label{sec:supp_raw}
We report the raw metrics behind the scores of Tabs.~\ref{tab:main} and~\ref{tab:ablation}, test the statistical significance of the comparison, and evaluate the sensitivity of InterMo to inference settings.

\begin{table}[t]
\centering
\caption{\textbf{Statistical significance.} Average-score gain of InterMo over each baseline with 95\% bootstrap intervals.}
\label{tab:bootstrap}
\small
\resizebox{\linewidth}{!}{\begin{tabular}{lccc}
\toprule
InterMo $-$ & Monologue & Conversation & All\\
\midrule
SemTalk & $+52.4$ {\scriptsize[47.2, 55.4]} & $+45.6$ {\scriptsize[35.0, 48.8]} & $+48.5$ {\scriptsize[39.2, 51.0]}\\
GestureLSM & $+7.9$ {\scriptsize[2.7, 11.0]} & $+13.0$ {\scriptsize[3.4, 16.2]} & $+13.7$ {\scriptsize[4.6, 16.7]}\\
DiffSHEG & $+28.0$ {\scriptsize[22.8, 30.7]} & $+20.5$ {\scriptsize[13.2, 24.5]} & $+23.4$ {\scriptsize[16.7, 27.1]}\\
MIBURI & $+13.6$ {\scriptsize[6.9, 19.2]} & $+23.0$ {\scriptsize[16.6, 26.4]} & $+21.7$ {\scriptsize[15.1, 24.7]}\\
Audio2Photoreal & $+22.9$ {\scriptsize[17.1, 25.6]} & $+26.0$ {\scriptsize[17.2, 28.4]} & $+28.1$ {\scriptsize[20.1, 29.5]}\\
DyaPlex & $+16.2$ {\scriptsize[8.9, 21.5]} & $+24.2$ {\scriptsize[14.6, 27.9]} & $+24.8$ {\scriptsize[15.5, 27.9]}\\
\bottomrule
\end{tabular}}
\end{table}

\begin{table}[t]
\centering
\caption{\textbf{Inference-time sensitivity of InterMo.} Voice-activity detector: turn-taking cues from the causal detector instead of the reference labels. History: motion/audio windows in steps. We report the average scores under different inference settings.}
\label{tab:sens}
\small
\resizebox{\linewidth}{!}{\begin{tabular}{lccc}
\toprule
Setting & Monologue & Conversation & All\\
\midrule
Default (CFG 4, 25/50) & 89.9 & 87.0 & 89.4\\
Seeds (mean$\pm$std, $n{=}3$) & 89.5$\pm$0.6 & 85.2$\pm$2.4 & 87.6$\pm$1.8\\
Voice-activity detector & 87.9 & 86.6 & 89.0\\
\midrule
CFG 1 & 44.1 & 50.1 & 48.3\\
CFG 2 & 62.1 & 72.9 & 70.1\\
CFG 3 & 74.3 & 85.4 & 85.0\\
CFG 5 & 88.0 & 77.4 & 80.5\\
CFG 6 & 84.3 & 72.0 & 74.7\\
\midrule
History $0.5\times$ (13/25) & 90.9 & 80.5 & 83.3\\
History $2\times$ (50/100) & 55.8 & 43.0 & 44.5\\
\bottomrule
\end{tabular}}
\end{table}

\begin{table}[t]
\centering
\caption{\textbf{Per-state arm speed.} Mean arm angular speed (deg/s) per conversational state.}
\label{tab:state_abs}
\small
\resizebox{\linewidth}{!}{\begin{tabular}{lcccc}
\toprule
Method & Speak & Listen & Pause & After partner\\
\midrule
GT & 37.8 & 15.9 & 28.8 & 20.1\\
InterMo & 44.4 & 19.5 & 34.3 & 19.0\\
GestureLSM & 34.9 & 19.7 & 29.2 & 20.2\\
DiffSHEG & 33.1 & 18.6 & 30.2 & 21.9\\
Audio2Photoreal & 26.5 & 23.7 & 24.9 & 22.5\\
MIBURI & 66.4 & 26.5 & 53.9 & 28.1\\
DyaPlex & 64.8 & 53.6 & 61.8 & 58.4\\
SemTalk & 7.4 & 6.2 & 7.4 & 6.6\\
\bottomrule
\end{tabular}}
\end{table}

\paragraph{Raw metrics of the comparison.}
Tabs.~\ref{tab:raw_solo} and~\ref{tab:raw_dyadic} report the raw metrics underlying Tab.~\ref{tab:main}.
DiffSHEG, SemTalk, and Audio2Photoreal have lower jerk than GT, which may reflect overly damped motion.
DyaPlex and MIBURI exceed GT in jerk, arm variance, and listening activity.

\begin{table*}[t]
\centering
\caption{\textbf{Raw metrics on the monologue subset.} GT values in the first row.}
\label{tab:raw_solo}
\small
\begin{tabular}{lccccccccc}
\toprule
Method & BC & FGD$\downarrow$ & FD-S$\downarrow$ & FD-K$\downarrow$ & AJ$\times10^3$ & PJ & AUJ & L1 arms & L1 torso\\
\midrule
GT & 0.565 & -- & -- & -- & 6.60 & 0.60 & 24.3 & 0.494 & 0.080\\
\midrule
SemTalk & 0.329 & 6.58 & 0.052 & 0.225 & 0.46 & 0.02 & 3.4 & 0.398 & 0.150\\
GestureLSM & 0.582 & 1.24 & 0.011 & 0.049 & 5.11 & 0.16 & 13.6 & 0.539 & 0.101\\
DiffSHEG & 0.428 & 1.13 & 0.031 & 0.019 & 1.71 & 0.11 & 5.0 & 0.705 & 0.101\\
MIBURI & 0.616 & 1.44 & 0.050 & 0.086 & 8.85 & 0.56 & 23.2 & 0.935 & 0.140\\
Audio2Photoreal & 0.492 & 2.74 & 0.023 & 0.039 & 2.56 & 1.22 & 7.9 & 0.560 & 0.091\\
DyaPlex & 0.641 & 2.30 & 0.032 & 0.063 & 10.42 & 0.56 & 29.0 & 0.733 & 0.114\\
InterMo (ours) & 0.558 & 0.86 & 0.017 & 0.016 & 6.35 & 0.43 & 15.5 & 0.418 & 0.085\\
\bottomrule
\end{tabular}
\vspace{8pt}
\caption{\textbf{Raw metrics on the conversation subset.} GT values in the first row.}
\label{tab:raw_dyadic}
\begin{tabular}{lccccccccccc}
\toprule
Method & BC & FGD$\downarrow$ & FD-S$\downarrow$ & FD-K$\downarrow$ & AJ$\times10^3$ & PJ & AUJ & L1 arms & L1 torso & LMA & ITA\\
\midrule
GT & 0.466 & -- & -- & -- & 6.30 & 0.68 & 23.2 & 0.406 & 0.068 & 1.00 & 0.036\\
\midrule
SemTalk & 0.288 & 4.94 & 0.036 & 0.101 & 0.46 & 0.02 & 3.5 & 0.351 & 0.108 & 0.67 & 0.001\\
GestureLSM & 0.548 & 0.99 & 0.009 & 0.007 & 4.93 & 0.14 & 12.4 & 0.561 & 0.087 & 1.37 & 0.015\\
DiffSHEG & 0.404 & 1.19 & 0.020 & 0.017 & 1.70 & 0.11 & 5.2 & 0.604 & 0.094 & 1.30 & 0.028\\
MIBURI & 0.581 & 2.12 & 0.102 & 0.130 & 9.59 & 0.89 & 33.0 & 1.048 & 0.166 & 1.94 & 0.052\\
Audio2Photoreal & 0.430 & 2.78 & 0.027 & 0.023 & 2.27 & 1.18 & 8.1 & 0.489 & 0.065 & 1.84 & 0.002\\
DyaPlex & 0.613 & 1.70 & 0.021 & 0.124 & 10.76 & 0.66 & 34.3 & 0.647 & 0.103 & 3.46 & 0.011\\
InterMo (ours) & 0.545 & 0.83 & 0.013 & 0.003 & 6.40 & 0.55 & 19.6 & 0.426 & 0.098 & 1.26 & 0.040\\
\bottomrule
\end{tabular}
\end{table*}

\paragraph{Raw metrics of the ablations.}
Tab.~\ref{tab:raw_abl} lists the raw values behind Tab.~\ref{tab:ablation}.
Removing partner speech also removes the partner entries of the turn-taking cues.
All ablated variants exceed GT in arm and torso--head L1 variance, which lowers their Diversity scores.
Removing interleaving, turn-taking cues, regularization, or mixed history also raises jerk clearly above GT.
Without mixed history, listening activity reaches 2.37 times the GT level and ITA drops to 0.015, which explains its Turn-taking score of 41.3.

\begin{table*}[t]
\centering
\caption{\textbf{Raw metrics of the ablations in Tab.~\ref{tab:ablation}.} GT values in the first row.}
\label{tab:raw_abl}
\small
\begin{tabular}{lccccccccccc}
\toprule
Configuration & BC & FGD$\downarrow$ & FD-S$\downarrow$ & FD-K$\downarrow$ & AJ$\times10^3$ & PJ & AUJ & L1 arms & L1 torso & LMA & ITA\\
\midrule
GT & 0.511 & -- & -- & -- & 6.40 & 0.65 & 23.6 & 0.433 & 0.071 & 1.00 & 0.036\\
\midrule
InterMo (full) & 0.551 & 0.68 & 0.012 & 0.003 & 6.38 & 0.51 & 18.3 & 0.424 & 0.094 & 1.26 & 0.040\\
w/o partner speech & 0.563 & 0.84 & 0.011 & 0.007 & 6.91 & 0.57 & 19.5 & 0.532 & 0.117 & 1.11 & 0.055\\
w/o interleaving & 0.587 & 0.71 & 0.021 & 0.012 & 8.04 & 0.64 & 24.8 & 0.500 & 0.107 & 1.75 & 0.030\\
w/o turn-taking cues & 0.571 & 0.89 & 0.012 & 0.006 & 7.61 & 0.57 & 22.2 & 0.525 & 0.116 & 1.50 & 0.048\\
w/o dynamics & 0.561 & 0.74 & 0.018 & 0.006 & 6.54 & 0.51 & 18.6 & 0.526 & 0.121 & 1.49 & 0.035\\
w/o regularization & 0.569 & 0.71 & 0.008 & 0.047 & 8.01 & 0.61 & 23.1 & 0.645 & 0.130 & 1.74 & 0.032\\
w/o mixed history & 0.599 & 0.95 & 0.011 & 0.042 & 9.10 & 0.73 & 28.7 & 0.615 & 0.138 & 2.37 & 0.015\\
\bottomrule
\end{tabular}
\end{table*}

\paragraph{Statistical significance.}
To test whether the comparison depends on the particular 200 clips, we draw 1{,}000 bootstrap resamples of the evaluation set.
The 64 monologue and 136 conversation clips are resampled separately with replacement, and all methods and GT share the same resamples.
All metrics, including GT references, are recomputed on each resample.
Tab.~\ref{tab:bootstrap} reports paired score differences between InterMo and each baseline with 95\% bootstrap intervals.
All intervals exclude zero, and InterMo scores higher in at least 99.8\% of resamples against every baseline on every subset.
The closest baseline is GestureLSM on monologues, with a gain of $+7.9$ and an interval of $[2.7, 11.0]$.

\paragraph{Inference-time sensitivity.}
Tab.~\ref{tab:sens} reports the overall score of InterMo under different inference settings, without retraining.
Across three sampling seeds, the overall score is $87.6{\pm}1.8$, and Tab.~\ref{tab:main} uses the first seed.
Low guidance scales of 1--2 under-use the speech condition and lose rhythm and diversity, while high scales of 5--6 exaggerate motion.
Scale 4 was fixed during development without a separate validation split, so this sweep is a post-hoc sensitivity analysis; it also gives the best overall score.
Halving the motion and audio windows (13/25 token steps) degrades the score moderately, whereas doubling them (50/100) beyond the training length collapses generation.
The voice-activity detector row recomputes only the cues fed to InterMo with the causal detector of Sec.~\ref{sec:supp_deploy}, keeping the GT references and metric labels unchanged.
On unscaled evaluation audio, this detector agrees with the reference labels on 83.6\% of self and 90.2\% of partner frames.
The average score drops from 89.4 to 89.0, while Turn-taking rises slightly from 84.5 to 86.2, within the seed variation.

\section{Interactive Motion Analysis}
\label{sec:supp_turns}
This section details the conversational-state analysis of Sec.~\ref{sec:turns}, with the state definitions and absolute speeds, and adds arm speed around turn events.
\begin{figure*}[t]
\centering
\includegraphics[width=\linewidth]{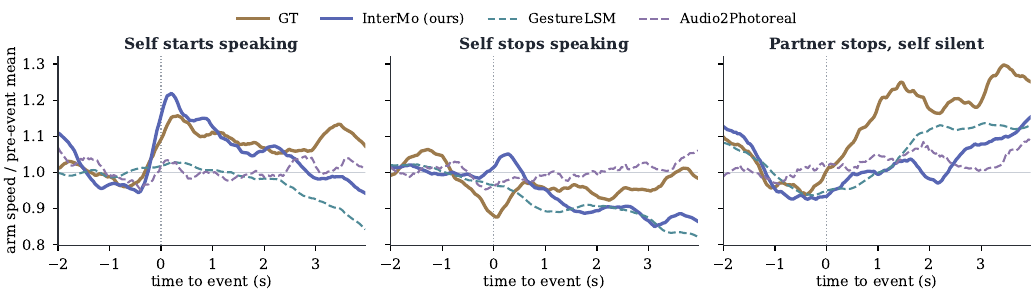}
\caption{\textbf{Arm speed around turn events.} Each curve is normalized by the method's mean over the 2\,s before the event.}
\label{fig:transition}
\end{figure*}

\paragraph{Per-state speeds.}
Frames of the 136 conversation clips are labeled from voice activity.
\emph{Speaking} frames have self activity, and \emph{listening} frames have partner activity while the target participant is silent.
On mutually silent frames, \emph{pause} means that the target participant spoke within the last 2\,s and more recently than the partner, and \emph{after partner} means the converse.
The remaining frames are long mutual silences and are not shown.
GT contains 31{,}622, 25{,}978, 7{,}108, and 5{,}650 frames of the four states.
Tab.~\ref{tab:state_abs} lists the absolute arm speeds underlying Fig.~\ref{fig:states}.
InterMo moves at approximately $1.2\times$ the GT speed while speaking, listening, and pausing, consistent with its LMA of 1.26, and at the GT level after the partner stops, so the contrasts between states are preserved.
GestureLSM and DiffSHEG match GT more closely in some states but compress the speaking--listening contrast.
MIBURI and DyaPlex move at about $1.7\times$ and $3.4\times$ the GT speed, respectively, while listening.

\paragraph{Turn transitions.}
We further align 733 self-speech starts, 742 self-speech ends, and 543 partner speech ends followed by self silence, and average the arm angular speed from $-2$ to $+4$\,s around them in Fig.~\ref{fig:transition}.
Like GT, InterMo shows an increase in arm speed at speech onset, while GestureLSM and Audio2Photoreal show little change. After the partner stops, GT speeds up within 2\,s, which InterMo and the baselines reproduce only partly.

\section{Real-Time Deployment}
\label{sec:supp_deploy}
This section describes the deployment settings, the inference speed and long-sequence stability, and the live demo and humanoid execution of Sec.~\ref{sec:applications}.

\paragraph{Deployment settings.}
In all deployments, each audio stream is scaled online to $-23$\,dBFS before entering the model.
A causal energy-based detector replaces the reference labels for the turn-taking cues.
A 40\,ms frame counts as speech when its energy exceeds $-40$\,dBFS, and each detection is extended by a 160\,ms hangover.
The detector reads the raw microphone audio in the live demo and the scaled audio otherwise.
A streaming self-collision correction is applied to the generated poses, whereas all other evaluations use the raw generator output.

\begin{figure*}[t]
\centering
\includegraphics[width=\textwidth]{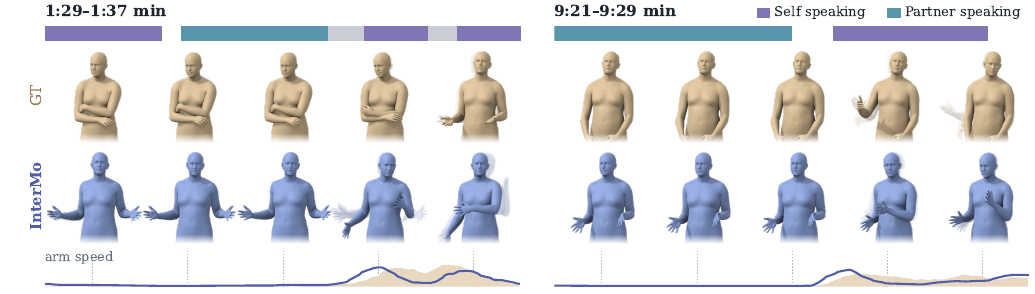}
\caption{\textbf{Long-sequence generation.} GT and InterMo are shown in early and late 8\,s windows of the same 10-minute conversation. InterMo generates the full sequence in one continuous stream. Bars indicate speech activity detected by the causal detector.}
\label{fig:supp_lseq}
\end{figure*}

\paragraph{Inference speed and stability.}
For inference speed, all timings use batch size 1 in fp32 on a single NVIDIA L20Z GPU and exclude model loading.
Tab.~\ref{tab:rtf} lists the time per 80\,ms step and the real-time factor of the generator and of the full deployment engine.
On the 200 evaluation clips, the generator takes 14.6\,ms per step on average and 19.3\,ms at the 95th percentile, and fewer than 0.4\% of steps exceed the 80\,ms budget.
For stability, the long-sequence run streams the first 10 minutes of 10 monologues and 23 conversations.
To check that turn-dependent motion does not fade over time, we compute the median speaking-to-listening arm-speed ratio in each 2.5-minute quarter of the conversations, which stays at 1.5--1.7 for InterMo against 1.6--2.1 for GT.
Fig.~\ref{fig:supp_lseq} shows GT and InterMo in an early and a late window of one conversation.
In both windows, InterMo stays still while the partner speaks and gestures when the target participant speaks, as GT does, and its arm speed relative to GT is unchanged from 1.5 to 9 minutes.

\begin{table}[t]
\centering
\caption{\textbf{Inference speed.} RTF is the generation time divided by the audio duration.}
\label{tab:rtf}
\small
\begin{tabular}{lcc}
\toprule
Setting & ms\,/\,step & RTF\\
\midrule
Generator & 14.6 & 0.19\\
Deployment engine & 26.0 & 0.32\\
\bottomrule
\end{tabular}
\end{table}

\begin{figure*}[t]
\centering
\includegraphics[width=\textwidth]{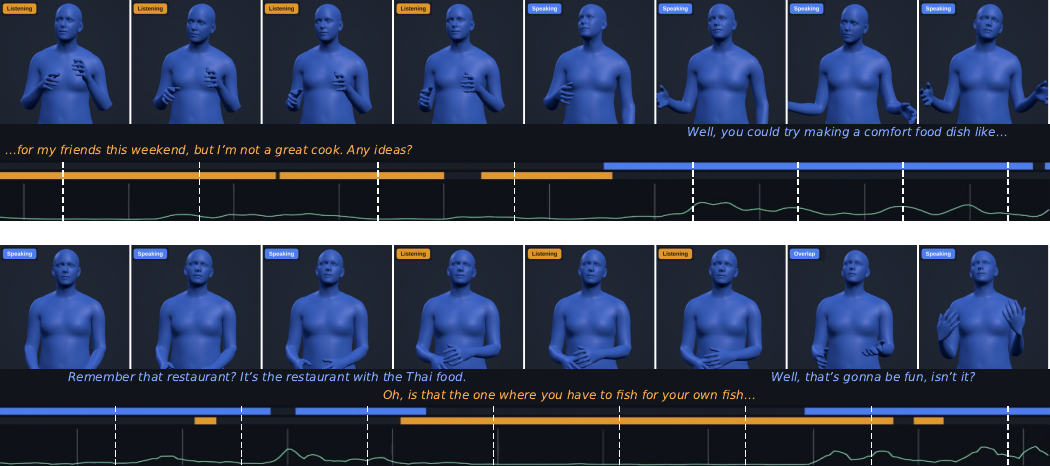}
\caption{\textbf{Live demo.} Top: dialogue mode, where the user's microphone is the partner stream and a spoken-dialogue model voices the avatar. Bottom: upload mode, where an uploaded speech track drives the avatar and an optional second track serves as the partner stream. 
Each row shows eight frames with their displayed state labels, followed by the transcript and timeline for the same 10\,s window. Blue marks the avatar's speech, orange the partner's speech, and the green curve the avatar's wrist speed. Dashed lines mark the frame times.}
\label{fig:supp_demo}
\end{figure*}

\begin{figure*}[t]
\centering
\includegraphics[width=\textwidth]{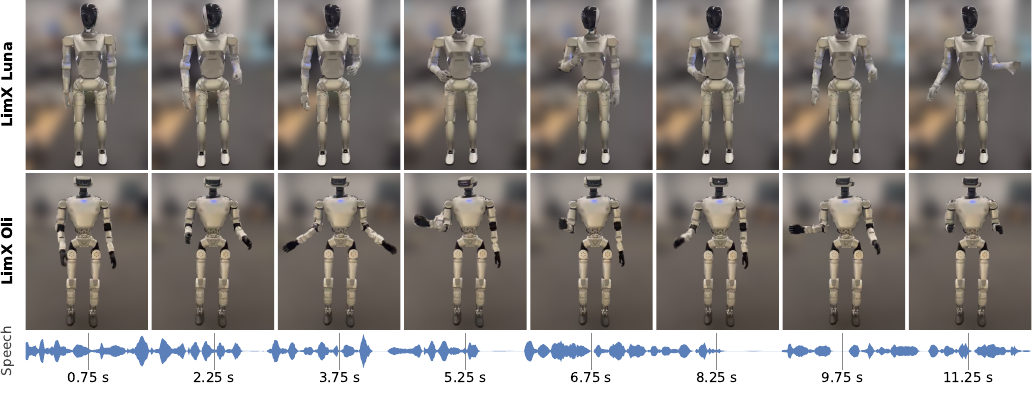}
\caption{\textbf{Humanoid execution from the same speech.} 
LimX Luna and LimX Oli are shown at 1.5\,s intervals over the first 11.25\,s of the input speech. The speech waveform appears below. Both recordings are aligned to the speech by audio cross-correlation.}
\label{fig:supp_robot}
\end{figure*}

\paragraph{Live demo and humanoid execution.}
The live demo drives a virtual avatar with our model in two modes, shown in Fig.~\ref{fig:supp_demo}.
In dialogue mode, the avatar converses with the user through a spoken-dialogue model, gesturing while it speaks and listening while the user speaks.
In upload mode, it generates motion from an uploaded speech track and an optional partner track.
Fig.~\ref{fig:supp_robot} shows LimX Luna~\cite{limx_luna} and LimX Oli~\cite{limx_oli} executing motions generated by InterMo from the same speech input, with both robots shown in full-body views. The same model runs on an onboard Jetson Orin NX without robot-specific training.

\section{Additional Qualitative Results}
\label{sec:supp_qual}
\begin{figure*}[t]
\centering
\includegraphics[width=\textwidth]{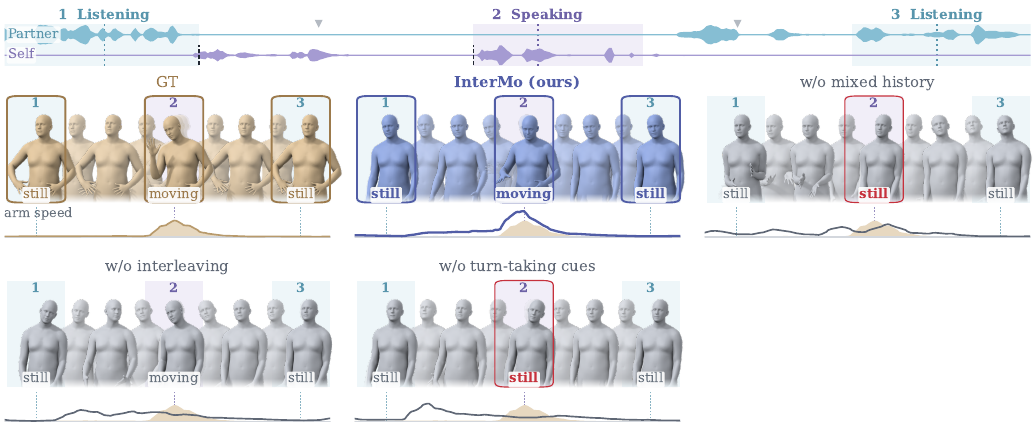}
\caption{\textbf{Qualitative ablation on the exchange of Fig.~\ref{fig:qualitative}.} The ablated variants miss the gesture after the target participant takes the turn or begin moving too early. Red boxes mark mismatches with GT.}
\label{fig:supp_ablation}
\end{figure*}

\begin{figure*}[t]
\centering
\includegraphics[width=\textwidth]{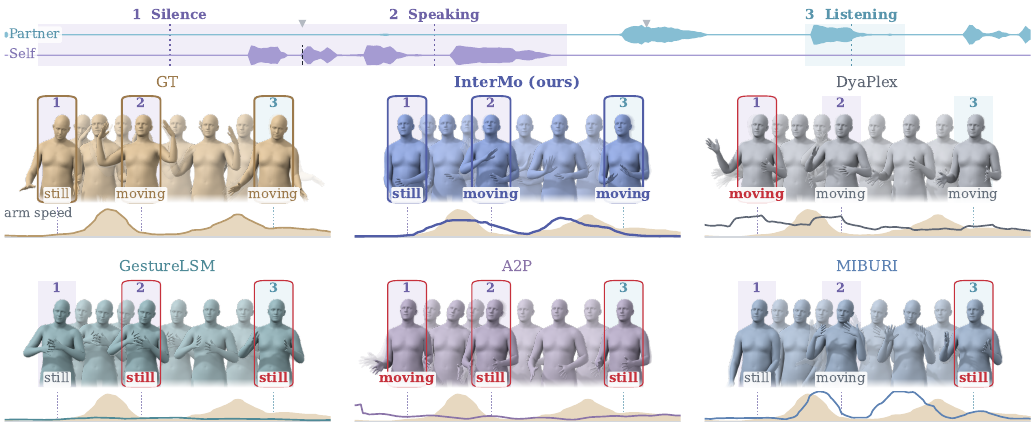}
\caption{\textbf{Additional qualitative comparison on a conversation.} InterMo remains still during silence, gestures while speaking, and continues moving while listening, consistent with GT. Red boxes mark mismatches with GT.}
\label{fig:supp_conv}
\end{figure*}

\begin{figure*}[t]
\centering
\includegraphics[width=\textwidth]{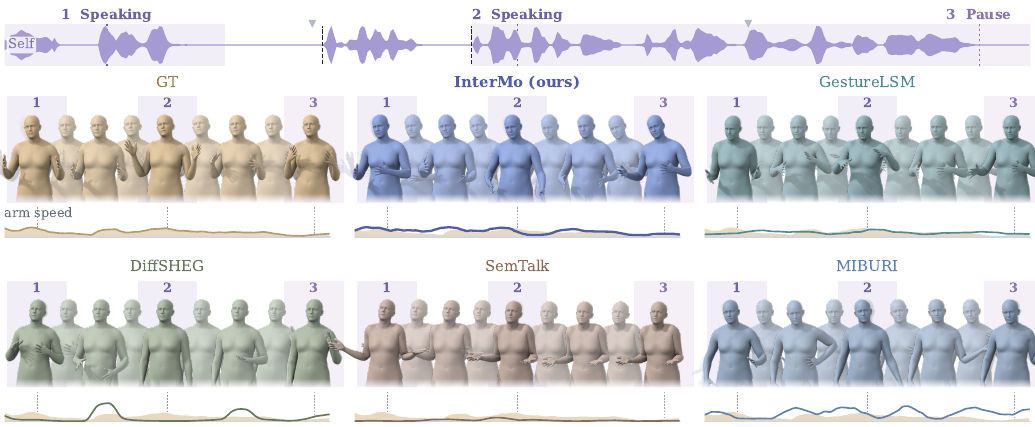}
\caption{\textbf{Additional qualitative comparison on a monologue.} InterMo gestures while speaking and settles at the pause as GT does.}
\label{fig:supp_mono}
\end{figure*}

Figs.~\ref{fig:supp_ablation}--\ref{fig:supp_mono} follow the layout of Fig.~\ref{fig:qualitative}, with Seamless Interaction clips for the conversation examples and a BEAT2 clip for the monologue example, all from our evaluation set.
A key moment is labeled moving when the mean elbow and wrist speed relative to the upper spine over the preceding 0.6\,s exceeds 0.15\,m/s, and still otherwise.
Red boxes mark key moments whose label differs from GT, and the plot beneath each method overlays its arm-speed curve with that of GT.
Fig.~\ref{fig:supp_mono} omits the still/moving labels and red boxes.

\paragraph{Qualitative ablations.}
Fig.~\ref{fig:supp_ablation} revisits the exchange of Fig.~\ref{fig:qualitative} with the ablated models of Tab.~\ref{tab:ablation}. Without mixed history or turn-taking cues, the model remains still when the target participant gestures after taking the turn, and without interleaving or turn-taking cues, it starts moving too early.

\paragraph{Conversation example.}
Fig.~\ref{fig:supp_conv} shows another conversation, in which the target participant is still during a silence, gestures while speaking, and keeps moving as the partner speaks.
InterMo agrees with GT at all three key moments, and each baseline misses at least one.

\paragraph{Monologue example.}
In Fig.~\ref{fig:supp_mono}, InterMo follows GT, gesturing at both speaking moments and settling at the pause.
GestureLSM, DiffSHEG, and SemTalk barely move during speech.
MIBURI keeps moving into the pause.

\end{document}